\documentclass[10pt,twocolumn,letterpaper]{article}

\makeatletter
\@namedef{ver@everyshi.sty}{}
\makeatother
\usepackage{cvpr}              

\usepackage{graphicx}
\usepackage{amsmath}
\usepackage{amssymb}
\usepackage{booktabs}
\usepackage[accsupp]{axessibility}
\usepackage[breaklinks,colorlinks]{hyperref}
\usepackage[capitalize]{cleveref} 
\usepackage[backref=true,doi=false,isbn=false,url=false,eprint=false,maxbibnames=10]{biblatex}
\usepackage[utf8]{inputenc} 
\usepackage[T1]{fontenc}    
\usepackage{dblfloatfix}    
\usepackage{url}            
\usepackage{booktabs}       
\usepackage{amsmath}        
\usepackage{amsfonts}       
\usepackage{nicefrac}       
\usepackage{xcolor}         
\usepackage{microtype}      
\usepackage{multirow}       
\usepackage{amssymb}        
\usepackage{pifont}         
\usepackage{makecell}       
\usepackage{tabularx}       
\usepackage{adjustbox}      
\usepackage{wrapfig}        
\usepackage{xspace}         
\usepackage{mwe}            
\usepackage{wrapfig}        
\usepackage{subcaption}     
\usepackage{csquotes}       
\usepackage[inline]{enumitem} 
\usepackage{placeins}       
\usepackage{tikz}           
\newcommand*\circled[1]{\tikz[baseline=(char.base)]{
            \node[shape=circle,draw,inner sep=1pt] (char) {#1};}}

\crefname{section}{Sec.}{Secs.}
\Crefname{section}{Section}{Sections}
\Crefname{table}{Table}{Tables}
\crefname{table}{Tab.}{Tabs.}

\DeclareMathOperator*{\argmax}{argmax}

\DeclareMathOperator*{\expectationop}{\mathbb{E}}
\newcommand{\expectation}[2]{\ensuremath{\expectationop_{#1}\left[#2\right]}}

\newcommand{\selectmodel}{\ensuremath{\phi_{\mathrm{select}}}}
\newcommand{\updatemodel}{\ensuremath{\phi_{\mathrm{update}}}}
\newcommand{\transformerselectmodel}{\ensuremath{\phi_\mathrm{select}^{\mathrm{tran}}}}
\newcommand{\linearselectmodel}{\ensuremath{\phi_\mathrm{select}^{\mathrm{linear}}}}
\newcommand{\threadembedding}{\ensuremath{\psi_\mathrm{thread}}}
\newcommand{\clipembedding}{\ensuremath{\psi_\mathrm{clip}}}
\newcommand{\tp}{\operatorname{TP}}
\newcommand{\fp}{\operatorname{FP}}
\newcommand{\tn}{\operatorname{TN}}
\newcommand{\fn}{\operatorname{FN}}
\definecolor{unred}{HTML}{D62728}
\definecolor{unblue}{HTML}{1F77B4}
\definecolor{ungreen}{HTML}{2CA02C}
\definecolor{unorange}{HTML}{FF7F0E}
\definecolor{unpurple}{HTML}{9467BD}
\definecolor{ungray}{HTML}{7f7f7f}

\crefname{appendix}{Appx.}{Appxs.}
\creflabelformat{equation}{#2\textup{#1}#3}

\DefineBibliographyStrings{english}{%
  backrefpage = {page},
  backrefpages = {pages},
}
\AtEveryBibitem{
  \clearfield{urlyear}
  \clearfield{urlmonth}
}
\setlist{nolistsep,leftmargin=*}

\makeatletter
\renewcommand{\paragraph}{%
  \@startsection{paragraph}%
                {4}
                {\z@}
                {1.2ex \@plus 0.5ex \@minus 0.5ex}
                {-1em}
                {\normalsize\bf}
}
\makeatother
\def\cvprPaperID{1811} 
\def\confName{CVPR}
\def\confYear{2022}

\begin{document}

\title{UnweaveNet: Unweaving Activity Stories}

\author{Will Price\\
University of Bristol\\
{\tt\small will.price@bristol.ac.uk}
\and
Carl Vondrick\\
Columbia University\\
{\tt\small vondrick.cs.columbia.edu}
\and
Dima Damen\\
University of Bristol\\
{\tt\small dima.damen@bristol.ac.uk}
}
\maketitle


\enlargethispage{-5cm}  
\noindent\begin{picture}(0,0)
\put(0,-433){\begin{minipage}{\textwidth}
\centering
\includegraphics[width=\linewidth]{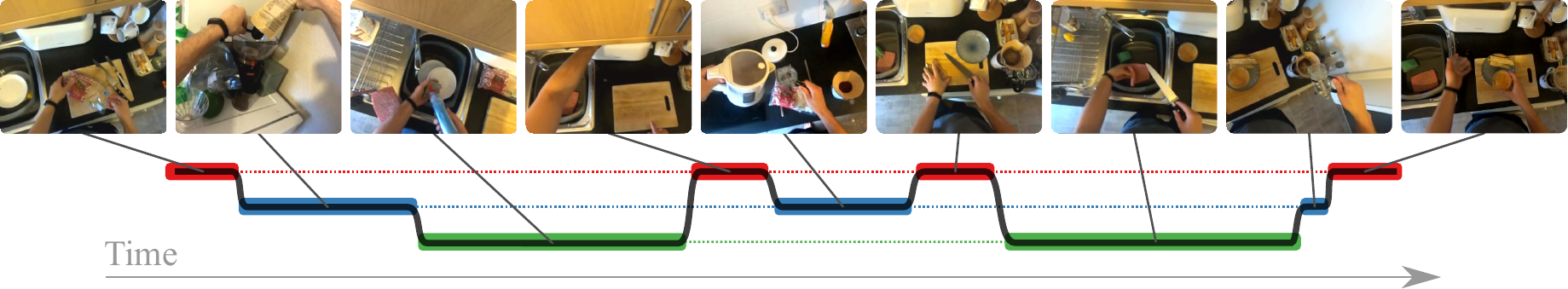}
\vspace{-2em}
\captionof{figure}{In our daily lives, one switches between activities (\eg{} \textcolor{unred}{making toast}, \textcolor{unblue}{preparing coffee}, \textcolor{ungreen}{washing up}) to minimize idle time.  Such behaviour results in video demonstrating multiple activities woven together. This paper introduces a model that learns to undo this, unweaving video into threads of activity without the need for semantic labels.}
\label{fig:unweaving-concept}
\end{minipage}}
\end{picture}%

\vspace{-10pt}
\begin{abstract}
  Our lives can be seen as a complex weaving of activities; we switch from one activity to another, to maximise our achievements or in reaction to demands placed upon us.
  Observing a video of unscripted daily activities, we parse the video into its constituent activity \textbf{threads} through a process we call \textbf{unweaving}.
  To accomplish this, we introduce a video representation explicitly capturing activity threads called a thread bank, along with a neural controller capable of detecting goal changes and resuming of past activities, together forming \textbf{UnweaveNet}.
  We train and evaluate UnweaveNet on sequences from the unscripted egocentric dataset EPIC-KITCHENS.
  We propose and showcase the efficacy of pretraining UnweaveNet in a self-supervised manner. 
\end{abstract}

\vspace{-20pt}
\section{Introduction}
 \begin{center}
 \begin{minipage}{1\linewidth}
   \textit{
``It's the morning and you've just walked into the kitchen: you're hungry, sleepy, the kitchen is a mess, but you have a paper to review for CVPR.
    You put some bread into the toaster, turn the kettle on to make coffee, and in between waiting for the kettle to boil and bread to toast, you clean the dishes.
    The toast pops up and you put it on a plate, then the kettle boils and you resume making your coffee, switching back and forth as necessary until your breakfast is ready.''
     }
   \vspace{0em}
 \end{minipage}
 \end{center}

As in the storyline described above and depicted in \cref{fig:unweaving-concept}, activities need not be completed over one continuous block of time.
Instead they are often paused and interleaved with other activities. 
This observation gives rise to a new interpretation of video as a \emph{weaving} of activities.
Such a perspective supports the distinction between two instances of an activity when the activity is paused and later resumed.
This distinction can be important for downstream applications, like assistive technologies which need to differentiate between starting a new task vs.\ resuming a previously paused one.

This novel view of video leads to the task proposed and tackled in this paper: \emph{unweaving} a video into its constituent activity threads.
Like a person reading a story mentally unweaves the story's narrative threads as they unfold, a model unweaving a video does so similarly, processing video \emph{online}, detecting new threads of activity as they appear and updating its representation of previously discovered threads as they are resumed.
Following this analogy, videos of activities as referred to as \emph{activity stories}.

\begin{figure}[t]
  \centering
  \includegraphics[width=1\linewidth]{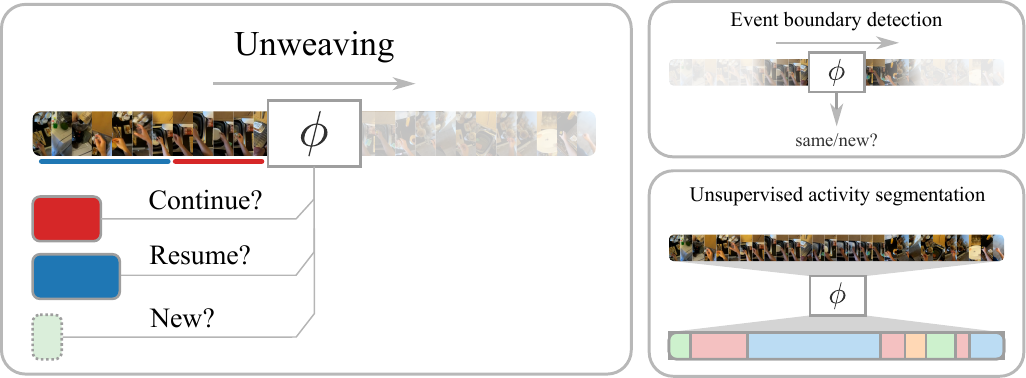}
  \caption{
        \textbf{Task comparison}: In \emph{unweaving}, the model has to decide, online, whether the current part of the video is a continuation of the last-seen activity, a resumption of previously paused activity, or a completely new instance of an activity.
    The figure compares unweaving to two previously studied tasks.
  }
  \label{unweavenet:fig:task-comparison}
  \vspace*{-14pt}
\end{figure}

This proposed task is related to two previously proposed tasks: event boundary detection and unsupervised activity segmentation.
The relationship between unweaving and these other related tasks is summarised in \cref{unweavenet:fig:task-comparison}.
Event boundary detection~\cite{shou2021_GenericEventBoundary,aakur2019_PerceptualPredictionFramework} aims to detect points in the video where a transition between two events occurs.
This task aims to model the experimental observation that humans can consistently detect transitions between events as they watch video online~\cite{zacks2001_Humanbrainactivity,zacks2001_Perceivingrememberingcommunicating,hard2006_Makingsenseabstract}.
Typically, these methods are performed online~\cite{shou2021_GenericEventBoundary,aakur2019_PerceptualPredictionFramework}, predicting the future video representation, comparing this against the true representation, and measuring the prediction error in order to decide whether a boundary can be detected.
\enlargethispage{-4.9cm} 
Compared to unweaving, event boundary detection focuses on finding the transitions between activities and doesn't support the association between events depicting a paused-and-resumed activity.
Also related, unsupervised activity segmentation~\cite{kukleva2019_UnsupervisedLearningAction,sarfraz2021_TemporallyWeightedHierarchicalClustering} clusters visual features to produce a segmentation of the video.
This task doesn't distinguish between different instances of the same activity, \eg{} the act of making two cups of tea, one after the other, is the same as making just one.
Unweaving is significantly more challenging as it is performed \emph{online}, without specifying the number of activities, nor the duration of the video.
Unweaving thus combines the challenges of the two aforementioned tasks.

In addition to introducing the problem of unweaving, this paper proposes a model that learns to unweave video into activity threads.
Different threads of activity are modelled in an explicit manner by a \emph{thread bank} that is manipulated by a neural controller as subsequent video is processed.
To train UnweaveNet, a self-supervised approach is introduced that leverages within-thread temporal-order consistency to construct \emph{synthetic} visual stories from unlabelled videos for pretraining.
The model is then finetuned using a small set of manually annotated stories.
The efficacy of this approach is shown experimentally using the unscripted egocentric dataset EPIC-KITCHENS-100~\cite{damen2020_RescalingEgocentricVision}.

Our contributions are summarised as:
\begin{enumerate*}[label=(\roman*)]
\item The novel task of unweaving video into its activity threads, online.
\item A new video representation explicitly modelling video as a set of activity threads operated by a neural controller, which together form \emph{UnweaveNet}.
\item A self-supervised pretraining approach for UnweaveNet that samples threads from different parts of a long video and synthetically forms woven activity stories.
\item Labelled annotations of activity threads from videos of the egocentric dataset: EPIC-KITCHENS\footnote{{\url{https://github.com/willprice/activity-stories}}}.
\item An empirical evaluation and ablation study of UnweaveNet.
\end{enumerate*}


\section{Related work}

\paragraph{Event boundary detection}
 In their seminal work, \textcite{zacks2001_Perceivingrememberingcommunicating} define an event as ``a segment of time at a given location that is perceived by an observer to have a beginning and an end''.
 \textcite{aakur2019_PerceptualPredictionFramework} propose a self-supervised method for detecting event boundaries, by predicting upcoming features.
A boundary is detected when the prediction error of the future frame exceeds a dynamically-set threshold.
\textcite{shou2021_GenericEventBoundary} introduce a new dataset for supervised event-boundary detection.
They explore detecting event boundaries using both supervised and unsupervised approaches. 
One of the unsupervised approaches, PredictAbility, measures the change in features about a point in time to detect boundaries.%

\paragraph{Action segmentation and detection}
In action \textit{segmentation}~\cite{huang2016_ConnectionistTemporalModeling,lea2017_TemporalConvolutionalNetworks,farha2019_MSTCNMultiStageTemporal,wang2020_Boundaryawarecascadenetworks,kukleva2019_UnsupervisedLearningAction,sarfraz2021_TemporallyWeightedHierarchicalClustering} the goal is to assign an action label to every frame.
In contrast, action \textit{detection}~\cite{sun2015_TemporalLocalizationFineGrained,shou2016_Temporalactionlocalization,shou2017_CDCConvolutionalDeConvolutionalNetworks,piergiovanni2019_TemporalGaussianMixture} predicts segments of video that possibly overlap. 
Most efforts for these tasks are supervised. 

\textcite{kukleva2019_UnsupervisedLearningAction} propose an unsupervised method for segmenting video by learning a  temporal embedding of frames.
First, they train an MLP to regress the position of a frame in the video from which it originates.
Intermediate features are extracted and act as the embedding of the frame.
The embeddings are then clustered using a constrained optimisation that prevents non-adjacent frames from being assigned to the same cluster.
An extension was proposed in \cite{vidalmata2021_Jointvisualtemporalembedding} that uses the embeddings from a model trained for future feature prediction.
\textcite{sarfraz2021_TemporallyWeightedHierarchicalClustering} also clusters frames, in an unsupervised manner, to form a temporally-weighted distance graph where nodes represent frames and edge weights are determined by the feature dissimilarity and temporal distance.
Frames are then clustered iteratively
until the desired number of clusters is reached.


\paragraph{Movie scene segmentation}
A variety of works tackle the problem of segmenting a movie into \textit{scenes}. 
All existing methods are offline and require specifying the number of scenes into which the movie will be split.
Early work by \textcite{yeung1995_Videobrowsingusing} introduced the concept of a hierarchical \textit{scene transition graph} which splits a movie into acts, scenes, and shots.
\textcite{cour2008_MovieScriptAlignment} use the screenplay and closed captions associated with a movie and introduce the problem of \textit{shot threading} to undo the common scenario in which shots from 2 or more cameras are interleaved together.
\textcite{tapaswi2014_StoryGraphsVisualizingCharacter} introduce a method for building a `StoryGraph', a type of visualisation, originally proposed by the web-comic xkcd~\cite{randallmunroe2009_MovieNarrativeCharts}, where each character in a TV episode is represented as a line on a 2D chart.

More recently, \textcite{rao2020_LocaltoGlobalApproachMultiModal} collect a large dataset MovieScenes containing 21k scene segments from 150 movies that are used to supervise their model.
These approaches are specific to movies which are made up of scenes and shots.
The notion of characters, multiple-cameras, shots, and scenes are not present in daily-activity videos.

\paragraph{Online clustering}
Unweaving videos can be viewed as a type of online clustering, where the number of clusters is not known ahead of time, nor the number of elements to be clustered.
\textcite{kulshreshtha2018_OnlineAlgorithmConstrained} investigate online clustering of faces in TV episodes.
Faces in the current shot are compared against those seen previously via patch feature similarity and are integrated into the existing closest cluster if the similarity exceeds a threshold.

In \textcite{damen_2014}, egocentric sequences from multiple users are used to cluster activities in an unsupervised manner, using 3D mapping and gaze information.

\textcite{nagarajan2020_EgoTopoEnvironmentAffordances} introduce a method for extracting a topological map of a kitchen environment from a first-person video.
Part of their method clusters contiguous portions of video into `activity-centric zones'.
To accomplish this, they train a Siamese network on pairs of video frames to determine whether the frames come from the same zone.
The network training is supervised using a heuristic: two frames are considered from the same zone if they are sufficiently close in time or if they have a shared background.
The map is constructed by processing the video sequentially, adding nodes and edges as new zones are discovered.
This method aims to aggregate subsequences of the video by location rather than by activity.

We compare to both online clustering and EGO-TOPO~\cite{nagarajan2020_EgoTopoEnvironmentAffordances} in our results.

\paragraph{Neural-network controlled machines}
There is extensive recent work on using neural networks to control data structures~\cite{grefenstette2015_Learningtransduceunbounded,joulin2015_Inferringalgorithmicpatterns,graves2014_NeuralTuringMachines,reed2016_NeuralProgrammerInterpreters,chen2020_CompositionalGeneralizationNeuralSymbolic,kurach2016_NeuralRandomAccessMachines,zaremba2016_LearningSimpleAlgorithms} such as neural stacks~\cite{grefenstette2015_Learningtransduceunbounded,joulin2015_Inferringalgorithmicpatterns}, neuro-symbolic stack machines~\cite{chen2020_CompositionalGeneralizationNeuralSymbolic}, and neural Turing Machines~\cite{graves2014_NeuralTuringMachines,zaremba2016_ReinforcementLearningNeural}.
UnweaveNet follows in this vein by using a neural controller to operate its thread bank.
Some of these works~\cite{grefenstette2015_Learningtransduceunbounded,graves2014_NeuralTuringMachines} use soft operations, where the model performs all operations simultaneously at each computation step with learnt weighting, whereas others~\cite{chen2020_CompositionalGeneralizationNeuralSymbolic} employ hard decisions as in UnweaveNet.
PtrNets~\cite{vinyals2015_PointerNetworks} introduced an approach for applying neural networks to seq2seq problems where the output sequence corresponds to locations in the input.
Part of UnweaveNet deals with a similar problem. 
UnweaveNet also shares similarities with the Memory Network~\cite{weston2015_MemoryNetworks}, however we use an adjustable memory size to support varying numbers of threads.


\paragraph{Video summarisation}
Another related task is video summarisation~\cite{zhang2016_VideoSummarizationLong,mahasseni2017_UnsupervisedVideoSummarization,rochan2018_VideoSummarizationUsing,yuan2019_CycleSUMCycleconsistentAdversarial,jung2019_DiscriminativeFeatureLearning}, which aims to extract highlights that give a condensed overview of the video.
Instead, in this work the full video is represented when unweaving an activity story into its constituent activities.


\section{Unweaving stories}
\label{chap:unweavenet:sec:method}
This section formulates the problem of unweaving (\cref{chap:unweavenet:sec:method:problem-description}); introduces the structured video representation (\cref{chap:unweavenet:sec:method:representation}) and the neural controller operating it  (\cref{chap:unweavenet:sec:method:controller}), together forming UnweaveNet; and concludes with the process used to train the model~(\cref{chap:unweavenet:sec:method:training}).
\begin{figure*}[t]
  \centering
  \includegraphics[width=.9\linewidth]{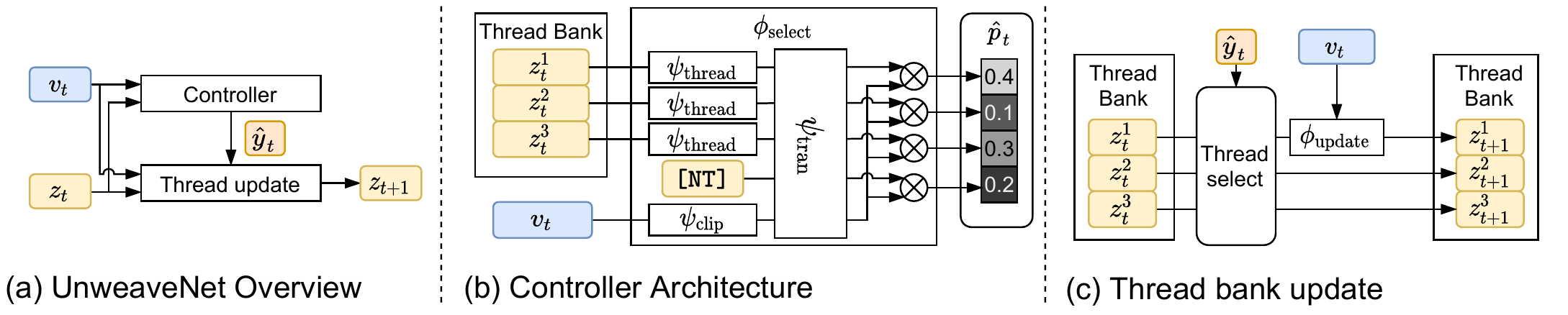}
  \vspace{-7pt}
  \caption[UnweaveNet architecture overview]{
  \begin{enumerate*}[label=\textbf{(\alph*)}]
    \item \textbf{UnweaveNet overview:} at each timestep, UnweaveNet receives a clip $v_t$ and considers the current thread bank state $z_{t}$; the controller determines whether that clip is a continuation of the ongoing thread, a resumption of a past thread, or the start of a new thread and updates the thread bank accordingly. 
    \item \textbf{Controller architecture:} $v_{t}$ and $z_{t}$ are embedded into a common space by linear layers $\clipembedding$ and $\threadembedding$ and fed into a transformer encoder. The clip embedding is compared against the thread embeddings and the new thread token \texttt{[NT]} to determine the probability of the clip joining an existing/new thread.
    \item \textbf{Thread bank update:} Given a thread to update, determined by $\hat{y}_t = \argmax_i p^i_t$, $\updatemodel$ incorporates $v_{t}$ into the current thread representation $z^i_t$ to produce the updated representation $z_{t+1}^i$.
    \end{enumerate*}
  }
  \vspace*{-12pt}
  \label{fig:architecture}
\end{figure*}

\subsection{Problem description}
\label{chap:unweavenet:sec:method:problem-description}
Unweaving is the problem of parsing an arbitrary-length video \emph{online} into $N$ variable-length activity threads, where $N$ is \emph{unspecified} and can vary across videos.
Once unwoven, all parts of the video belonging to the same thread should correspond to one activity instance.
When the video portrays a switch to a different activity, the ongoing thread should be paused and a different thread started or resumed. 
Assuming $\hat n_{t}$ threads have been identified up to time $t$, the task is to decide whether the current video clip $v_t$ is a continuation of an existing thread or the beginning of a new thread.\footnote{The clip is assumed short enough to belong to one thread, leaving the exploration of clip length and multi-thread clips to future work.}

\subsection{Thread bank}
\label{chap:unweavenet:sec:method:representation}

Core to our proposal is a structured representation of video, which we call a \emph{thread bank}. This stores the representations of all complete and on-going activity threads discovered in the video as it is processed.
New activity threads can be added into the bank as they are discovered and existing threads can be updated by incorporating new clips into them.
In its most general form, a representation $z^i_t$ of thread $i$ at time $t$ is produced as an aggregation $g$ of the set of clips $\mathcal{V}_t^i$ currently assigned to the thread:
\begin{equation}
    z_{t}^i = g\left(\mathcal{V}^i_{t}\right),\qquad g : \mathbb{R}^{\left|\mathcal{V}^i_t\right| \times C} \rightarrow \mathbb{R}^D
    \label{eq:update-model}
\end{equation}
where $C$ is the dimension of the clip feature and $D$ the dimension of the thread representation.
However, this doesn't quite capture the concept of an activity as an evolving process.
Instead, a recurrent function $\updatemodel$ is used in place of $g$, better modelling this perspective by updating the activity representation with information from the latest clip:
\begin{equation}
    z_{t+1}^i = \updatemodel(v_{t}, z_t^i),\qquad \updatemodel : \mathbb{R}^C\times \mathbb{R}^D \rightarrow \mathbb{R}^D.
    \label{eq:thread-update-recurrent}
\end{equation}
When a new thread is discovered, $z^i_t$ is replaced with an initial learnt empty-thread representation $z^*$.

The state of the thread bank at time $t$ and $t+1$ can be related as follows.
Let $\hat{y}_{t}$ be the thread to which $v_{t}$ will be added; for UnweaveNet, this is decided by its neural controller (described in \cref{chap:unweavenet:sec:method:controller}).
The representations within the updated thread bank $z_{t+1}$ are related to the previous previous representations $z_t$ as follows
\begin{equation}
  \label{eq:thread-update}
  z_{t+1}^i = \begin{cases}
    \updatemodel(v_{t}, z_t^i) &  i = \hat{y}_{t} \leq \hat n_t\\
    \updatemodel(v_{t}, z^*) & i = \hat{y}_{t} = \hat n_t + 1\\
    z_t^i &  \mathrm{otherwise.}
   \end{cases}
\end{equation}
When $t = 1$, the thread bank is empty, thus $\hat n_1 = 0$.

While the number of threads in the bank can vary, each thread's representation is fixed in size, thus the model's complexity is linear in the number of threads rather than number of clips.
Since the number of clips greatly exceeds the number of threads, this keeps the representation compact.

\subsection{Neural controller}
\label{chap:unweavenet:sec:method:controller}
In order to construct the thread bank representation, a neural controller manipulates the thread bank as new clips are considered from the video~(\cref{fig:architecture}a).
Given a new clip, the controller determines whether the clip is the beginning of a new thread or whether it is a continuation or a resumption of an existing thread (\cref{fig:architecture}b).
Once the decision has been made, the thread bank is updated (\cref{fig:architecture}c) and the process iterates.

UnweaveNet uses a neural network $\selectmodel$ to implement the controller.
It is fed the new clip $v_{t}$ and the current thread bank state $z_t$ and is tasked with calculating the probabilities $p_{t} \in [0,1]^{\hat n_t + 1}$ of how likely it is that $v_{t}$ is the continuation of the ongoing thread, the resumption of a past thread, or the start of a new thread.
Specifically, $p^{1:\hat n_t}_t$ contains the probabilities of $v_t$ continuing/resuming existing threads
and $p^{\hat n_t + 1}_t$ is the probability of $v_t$ starting a new thread.
The controller $\selectmodel$ computes a vector of logits 
\begin{equation}
  l_t = \selectmodel(v_{t}, z_{t}), \quad
  \selectmodel : \mathbb{R}^C \times \mathbb{R}^{\hat n_t \times D} \rightarrow \mathbb{R}^{\hat n_{t} + 1},
  \label{eq:select-model}
\end{equation}
which are softmaxed (with temperature $\tau$) to compute~$p_t$
\begin{equation}
  p^i_{t}  = e^{l^i_t/\tau}/ \textstyle \sum_{j=1}^{\hat n_t + 1}e^{l^j_t/\tau}.
  \label{eq:probs}
\end{equation}
\vspace*{-12pt}

\noindent The decision is then determined by $\hat y_t = \argmax_i p^i_t$.

To obtain $l^{1:\hat n_t}_t$, we learn a space in which the clip is closest to the thread it belongs to.
Both the clip and threads are embedded into this space through linear projections $\clipembedding: \mathbb{R}^{C} \rightarrow \mathbb{R}^E$ and $\threadembedding: \mathbb{R}^D \rightarrow \mathbb{R}^E$.
The cosine similarity between the clip and each thread embedding is measured to produce the scores $l_t^{1:\hat n_t}$ for how likely it is that the clip belongs to each thread.
We also learn a latent similarity score $l^\mathrm{NT} \in \mathbb{R}$, which acts as a threshold that a clip-thread similarity must exceed if the clip is to be deemed a continuation.
This gives rise to the linear controller
\begin{equation}
\resizebox{0.98\linewidth}{!}{%
$\linearselectmodel\left(v_t, z_t\right)_i = \begin{cases}
    \cos\left(\clipembedding\left(v_t\right), \threadembedding\left(z_t^i\right)\right) & i \leq \hat n_t \\
    l^\mathrm{NT} & \mathrm{otherwise}.
  \end{cases}%
$%
}\end{equation}


\noindent We contrast this approach with a contextualised approach
by feeding a transformer encoder $\psi_\mathrm{tran}$ with a sequence composed of the embedded clip and thread representations, and a new-thread token $\mathtt{[NT]} \in \mathbb{R}^E$.
The transformer $\psi_\mathrm{tran}$ takes this input sequence
$  \left[ \clipembedding\left(v_t \right), \threadembedding\left(z^1_t\right), \ldots, \threadembedding\left(z^{\hat n_t}_1\right), \mathtt{[NT]} \right] $
and produces a corresponding output sequence
  $\left[\tilde v_t, \tilde z^1_t, \ldots, \tilde z^{\hat n_t + 1}_t \right]$. 
These contextualised vectors (denoted by tilde) form a new embedding space in which the clip $\tilde v_t$ is compared against all threads $\tilde z_t^{1:\hat n_t}$ \emph{and} the new-thread token $\tilde z^{\hat n_t +1}_t$to form the vector of logits:
\begin{equation}
  \transformerselectmodel(v_t, z_t)_i = \cos(\tilde{v}_t, \tilde{z}^i_t).
\end{equation}
This process using $\transformerselectmodel$ is graphically depicted in \cref{fig:architecture}b.

\begin{figure}[t]
  \centering
  \includegraphics[width=.5\linewidth]{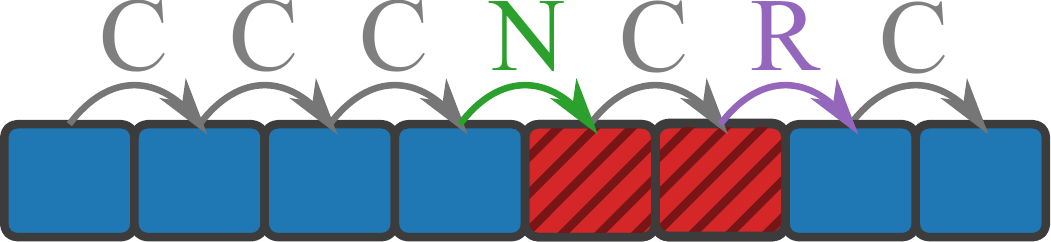}
  \caption[Decision decision scenarios faced by UnweaveNet during the unweaving process]{
    \textbf{Different decision scenarios during the unweaving process}: \textcolor{ungray}{C}ontinue thread, \textcolor{ungreen}{N}ew thread, \textcolor{unpurple}{R}esume thread.
    Each square depicts a single clip which is coloured by the thread it belongs to.
  }
  \vspace*{-12pt}
  \label{fig:decision-scenarios}
\end{figure}

\begin{figure*}[t]
  \centering
  \includegraphics[width=.8\linewidth]{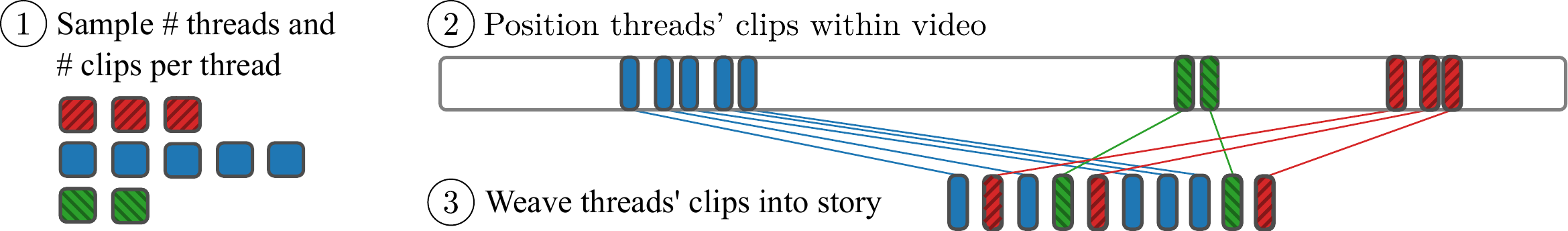}
  \caption[The synthetic story construction process]{
    \textbf{Synthetic story construction:}
    \circled{1} The number of threads is sampled, then the quantity of clips comprising each thread are sampled.
    \circled{2} The threads are randomly positioned within the video, where clips within a thread are separated by a small random gap.
    \circled{3} Finally, the threads' clips are randomly woven together into a synthetic story.
  }
  \vspace*{-12pt}
  \label{fig:synthetic-story-formation}
\end{figure*}

\subsection{Training}
\label{chap:unweavenet:sec:method:training}
UnweaveNet is trained \emph{end-to-end}, including the backbone used to extract clip features, thus the clip and thread representations are optimised along with the controller parameters.
The decisions made by $\selectmodel$ are supervised using teacher forcing~\cite{williams1989_LearningAlgorithmContinually,goodfellow2016_DeepLearning}, used for training language models;
at each time step, $z_t$ is populated according to the ground-truth clip-thread assignments $y_{1:{t-1}}$.
A loss is then imposed on the output $p_{t}$ (\cref{eq:select-model}) with the correct decision~$y_{t}$.

Due to the imbalance in the decisions, we weight three mutually exclusive scenarios in the loss (\cref{fig:decision-scenarios}): starting a new thread (N), continuing the currently-active thread (C), and resuming a paused thread after a gap of more than one clip (R).
Each scenario $s \in \{\mathrm{C,N,R}\}$ is given a positive weight $\alpha_{s}$, and we train with a focal loss~\cite{lin2017_FocalLossDense} that causes hard examples to have a larger impact on the gradient than easy examples.
Let $S$ be a function that given $y_{1:t}$ determines the scenario $s$ at time $t$.
The loss (with focal hyperparameter~$\gamma$) for a single story takes the form:
\begin{equation}
  \label{eq:loss}
  \mathcal{L} = -\sum_{t} \alpha_{S(y_{1:t})}\left(1 - p_t^{y_t}\right)^\gamma \log{p^{y_t}_t}.
\end{equation}
The loss is averaged over all stories within the batch and back-propagated to train UnweaveNet.


This section formalised the problem of video unweaving (\cref{chap:unweavenet:sec:method:problem-description}) and proposed a model, UnweaveNet, for solving it.
UnweaveNet builds up a structured representation of video, called a thread bank, as it processes a streaming video (\cref{chap:unweavenet:sec:method:representation}).
A neural controller (\cref{chap:unweavenet:sec:method:controller}) determines whether a clip belongs to an existing thread, or is the beginning of a new, and updates the thread bank accordingly.
UnweaveNet is trained through a teacher-forcing set-up (\cref{chap:unweavenet:sec:method:training}).

\section{Obtaining stories}
\label{chap:unweavenet:sec:datasets}

Exploring unweaving requires a dataset of untrimmed videos with interleaved activity instances.
We use the large-scale unscripted egocentric dataset EPIC-KITCHENS~\cite{damen2020_RescalingEgocentricVision}, where 
videos capture people in their own kitchens, capturing their activities over a three day period using a head-mounted camera.
The dataset contains videos of participants switching back and forth between activities, making it a suitable source for obtaining interesting activity stories to \emph{unweave}.

First, the untrimmed nature of the dataset is leveraged in making \emph{synthetic stories}, that can be acquired \emph{without} any annotations for pretraining UnweaveNet.
Second, a sample of the dataset is annotated with activity threads providing \emph{activity stories} for finetuning and evaluation purposes.

\paragraph{Synthetic stories}
\label{chap:unweavenet:appendix:synthetic-story-construction}


We propose a method for pretraining UnweaveNet in a self-supervised manner by constructing synthetic stories through a randomised sampling procedure applied to long video.
We sample threads from the same video as threads from different videos would be trivial to unweave.
Given a number of threads $N$, we sample $N$ sequences of clips of varying lengths, each from distinct locations within the same video, to produce synthetic threads.

Similar to the assumption in instance discrimination~\cite{chen2020_SimpleFrameworkContrastive,chen2020_ImprovedBaselinesMomentum}, we assume these threads depict distinct activities.
To further increase the likelihood of this, we enforce a minimum separation between threads.
We then randomly interleave these together, respecting the arrow-of-time of the clips within each thread, to produce a synthetic story (depicted in \cref{fig:synthetic-story-formation}).
Full details on this sampling process are given in \cref{appendix:synthetic-stories}.

In our setup, each synthetic story is composed of 10 clips woven from 1--4 threads.
Synthetic stories are sampled randomly, per batch, from the dataset's training videos.
Thus, the model is trained on a practically infinite number of synthetic stories.
In all experiments, we sample 800K synthetic stories (8M clips, 50k batches, each containing 16 stories).


\paragraph{Activity-story annotation}
Synthetic stories contain visual discontinuities and synthetic threads aren't always composed of clips from a single activity due to the random sampling process.
Thus, a model trained solely on synthetic stories falls short of being able to unweave natural video into activity threads.
Consequently, a small dataset of manually annotated activity stories was collected for finetuning and evaluation purposes (further details are given in \cref{appendix:activity-stories}).


Overall, 15k clips were annotated  (4.2 hours) across 448 videos from EPIC-KITCHENS into their activity threads.
Of these clips, 9.5k are for training, 3.8k for validation, and 1.8k for testing.
The activity stories comprising the training and validation set were collected by 7 volunteer annotators, all consisting of 10 clips.
\setlength{\wrapoverhang}{-10pt} 
\begin{wraptable}[12]{r}{.4\linewidth}
  \vspace{-5pt}
  \begin{tabular}{@{}lrrr@{}}
    \toprule
    & \multicolumn{3}{c}{\# Threads} \\
    \cmidrule{2-4}
    Split & 1      & 2      & 3    \\
    \midrule
    Train & 718    & 201    & 32   \\
    Val   & 211    & 94     & 46   \\
    Test  & 50     & 50     & 50   \\
    Total & 979    &345     & 128\\
    \bottomrule
  \end{tabular}
  \caption{EPIC-KITCHENS activity-story dataset by \# of threads.}
  \label{tab:activity-dataset-stats}
\end{wraptable}
A sample of each annotator's stories were checked for correctness.
For testing, we manually collected stories of varying lengths (from 5 to 26 clips).
The training stories come from videos in the training split of the EPIC-KITCHENS action-recognition challenge, and the test and validation stories come from videos from the combined test and validation splits.
Statistics on the number of stories by number of threads are given in \cref{tab:activity-dataset-stats}.
Note that \emph{threads are not annotated with any semantic labels}, the only metadata annotated is which clips belong together within a thread.

\section{Experiments}
\label{chap:unweavenet:sec:experiments}

This section evaluates UnweaveNet on the EPIC-KITCHENS activity-story dataset, demonstrating the model's capabilities both qualitatively and quantitatively.
Its performance is compared to a number of baselines and design decisions are motivated through ablation studies.
The section is structured as follows:
\cref{chap:unweavenet:baselines} introduces the baselines; \cref{chap:unweavenet:metrics} explains the evaluation metrics; \cref{chap:unweavenet:sec:results} presents the main results; and \cref{chap:unweavenet:sec:ablation-studies} ablates the components in UnweaveNet.

\subsection{Experimental setup \& Baselines}
\label{chap:unweavenet:baselines}

This section briefly outlines the key points of the experimental setup (comprehensive details can be found in \cref{appendix:experimental-details}) and explains the different baselines.
As unweaving is a new concept, there are no existing works to directly compare against.
Accordingly, a variety of baselines are either proposed or adapted from prior work.
Two non-learnt na\"ive baselines are provided to give a lower bound on performance.
As unweaving is an inherently online process, methods for offline unsupervised action segmentation are excluded.

\paragraph{Experimental setup}
Videos are encoded to 16FPS, resized to a height of 112px, and center-cropped.
Horizontal flipping is used during training for data augmentation.
The backbone network used to extract clip features is a top-heavy 3D ResNet-18 pretrained on Kinetics~\cite{kay2017_KineticsHumanAction} using the DPC self-supervised objective~\cite{han2019_VideoRepresentationLearning}.
A single-layer GRU~\cite{cho-al-emnlp14} is used as $\updatemodel$ and a single-layer transformer encoder with 4 heads is used for $\transformerselectmodel$.

\paragraph{Na\"ive baselines}
The simplest baseline assigns all clips to a single thread, and hence is referred to as the \emph{single-thread} baseline.
Naturally this baseline will perform optimally in the case of all clips belonging to the same thread. 
An alternate, non-learnt baseline predicts all possible partitions of the clips as equally likely, including new threads.
This is termed the \emph{random-chance} baseline.

\paragraph{Online clustering}
This baseline clusters clips online into threads by measuring feature similarities.
The similarity $s^i_t$ of clip $v_t$ to each thread $\mathcal{V}^i_t$ detected up to time $t$ is
\begin{equation}
  s^i_t = \frac{1}{\left|\mathcal{V}^i_t\right|} \sum_{v_j \in \mathcal{V}^i_t} \cos\left(v_t, v_j\right)
\end{equation}
The clip is assigned a candidate thread $\argmax_i s^i_t$ to join and if $\max_i s^i_t$ is beyond a specified threshold, then the clip continues that thread, otherwise a new thread consisting only of the clip $v_t$ is started. The threshold is trained optimally on the validation set.

\paragraph{PredictAbility~\cite{shou2021_GenericEventBoundary}}
This is a model designed for event segmentation adapted to perform unweaving.
Recall that event segmentation aims to find the boundaries between events. 
This model detects event boundaries where there is a large change in feature representation over time.
These boundaries are snapped to clip edges which splits the video into threads (without any resumed threads).

\paragraph{EGO-TOPO~\cite{nagarajan2020_EgoTopoEnvironmentAffordances}}
This is a model designed to produce a graph structure of the egocentric video, where subsequences captured in the same physical location comprise a node.
Edges are formed when the video depicts a transition from one location to another.
Thus EGO-TOPO models the video by the topological locations depicted within.
Nevertheless, it can still be viewed as a form of video unweaving that creates threads by location.
As this model operates on frames rather than clips, a majority voting strategy is used to map from the frames assigned to nodes in the graph to the clips comprising threads in the unweaving.

\subsection{Metrics}
\label{chap:unweavenet:metrics}

To measure performance on the unweaving task, we report the following metrics (further details, with metric equations, are given in \cref{appendix:metrics}):
\vspace*{6pt}

\begin{itemize}
\item The \emph{Rand Index (RI)}~\cite{rand1971_ObjectiveCriteriaEvaluation}, often used in clustering problems, which compares the estimated pair-wise grouping of clips to the ground truth.
\item The \emph{Teacher-Forcing Accuracy (TFA)}, which reports accuracy of the decisions produced by $\selectmodel$ at each timestep $t$ when we populate the thread bank according to the ground truth $y_{1:t-1}$.
  This allows us to evaluate the model's performance at each timestep without confounding the results by erroneous past decisions.
\item $\Delta N$, the difference in the number of predicted threads to the ground truth. This allows us to compare methods on whether new threads are created too readily, or too infrequently.
\end{itemize}


\subsection{Results}
\label{chap:unweavenet:sec:results}

\begin{table*}[t]
  \centering
  \resizebox{\linewidth}{!}{
  \begin{tabular}{ll rrrr rrrr rrrr}
    \toprule
                   &                   & \multicolumn{4}{c}{RI ($\uparrow$)}                                 & \multicolumn{4}{c}{TFA ($\uparrow$)}                                       & \multicolumn{4}{c}{$\Delta N (\rightarrow 0 \leftarrow)$} \\

                                       \cmidrule(lr){3-6}                                                          \cmidrule(l){7-10}                                                          \cmidrule(l){11-14}
    Model          & Supervision       & 1                & 2                & 3                & Avg              & 1                & 2                & 3                & Avg              & 1               & 2               & 3                & Avg \\
    \midrule
    Single thread  & -                 & -                & 52.4             & 35.3             & 62.6             & -                & 41.9             & 26.5             & 56.2             & -               & -1.0            & -2.0             & -1.0 \\
    Chance$^*$     & -                 & 18.9             & 48.4             & 60.5             & 42.6             & 50.0             & 36.0             & 26.7             & 37.6             & 3.7             & 3.0             & 3.8              & 3.5 \\
    \cmidrule{1-14}
    PredictAbility~\cite{shou2021_GenericEventBoundary} \hspace{12pt}
                   & Self-supervised   & 47.2             & 58.8             & 73.5             & 59.9             & 54.6             & 44.4             & 39.3             & 46.1             & 1.2             & {\bfseries 0.5} & 0.5              & 0.8 \\
    Online Clustering
                   & Self-supervised   & 66.8             & 60.4             & 64.3             & 64.0             & {\bfseries 92.4} & 60.5             & 47.5             & 63.7             & 1.0               & 0.9              & 0.3              & 0.7 \\
    EGO-TOPO~\cite{nagarajan2020_EgoTopoEnvironmentAffordances}
                   & Action boundaries\hspace{12pt} & 83.9             & 66.2             & 64.9             & 71.7             & -                & -                & -                & -                & 0.6             & 0.7             & 1.6              & 0.9 \\
    \cmidrule{1-14}
    UnweaveNet     & SS                & 63.8             & 68.6             & {\bfseries 74.8} & 69.1             & 78.3             & 53.1             & 61.9             & 66.5             & 1.5             & 2.4             & 2.0              & 2.0 \\
    UnweaveNet     & AS                & 81.8             & 58.4             & 60.9             & 67.0             & 83.0             & 51.3             & 37.6             & 57.3             & 0.6             & {\bfseries 0.5} & {\bfseries -0.1} & {\bfseries 0.3} \\
    UnweaveNet     & SS+AS             & {\bfseries 84.1} & {\bfseries 70.1} & 71.3             & {\bfseries 75.1} & 85.2             & {\bfseries 71.9} & {\bfseries 74.3} & {\bfseries 77.5} & 0.5             & 0.7             & 0.3              & 0.5 \\
    \bottomrule
  \end{tabular}}
  \vspace{-7pt}
  \caption[Quantitative evaluation of UnweaveNet on the EPIC-KITCHENS activity-story test set]{
  Unweaving performance (5 run average) on the activity-story test set for na\"ive no-learning baselines (top section), learnt baselines (middle section) and UnweaveNet (bottom section) with/without pretraining on synthetic stories~(SS) and finetuning on activity stories~(AS).
  Metrics are described in \cref{chap:unweavenet:metrics}.
    Performance is broken down by the number of threads in the test story (specified below each metric heading).
    Chance$^*$ refers to a random partition for RI and $\Delta N$, and a random decision at each step for teacher-forcing accuracy (TFA).
    }
  \label{tab:performance}
\end{table*}

The results for unweaving the stories from the EPIC-KITCHENS activity-story test set are presented in \cref{tab:performance}, comparing the non-learnt and learnt baselines to UnweaveNet.
UnweaveNet's performance is reported under three different training regimes: synthetic stories only (SS), activity stories only (AS), and pretrained on synthetic stories then finetuned on the activity stories (SS+AS).
UnweaveNet performs well compared to the baselines.
The full model that is pretrained on synthetic stories and finetuned on activity stories (SS+AS) outperforms the baselines on all averaged metrics.
EGO-TOPO performs best out of the baselines, however this is primarily due to its strong performance on the single-thread examples which demonstrate a sole activity, typically in one location\footnote{It is not possible to evaluate the TFA on the EGO-TOPO model with the provided implementation.}.
Compared to UnweaveNet (when trained with activity stories), all the learnt baselines have a tendency to create more threads than exist in the ground truth as evidenced by the positive $\Delta N$ values.
UnweaveNet also suffers this issue, albeit to a lesser extent.

The benefit of synthetic story pretraining is evident comparing AS to SS+AS under both the RI and TFA metrics.
The average $\Delta N$ is closer to the ideal (0) for the AS model compared to the SS+AS model.
This can be attributed to the synthetic stories having more threads than the activity stories, thus predisposing the model to creating more threads.
Interestingly, the average RI is higher, 69.1\% vs.\ 67.0\%, when \emph{only} pretraining on synthetic stories (SS) than when only training on activity stories (AS); though the SS model oversegments the video into too many threads as can be seen from the higher average $\Delta N$ (2.0 vs.\ 0.3) and higher 3-thread RI (74.8\% vs.\ 60.9\%).

\begin{figure}[t]
    \includegraphics[width=\linewidth]{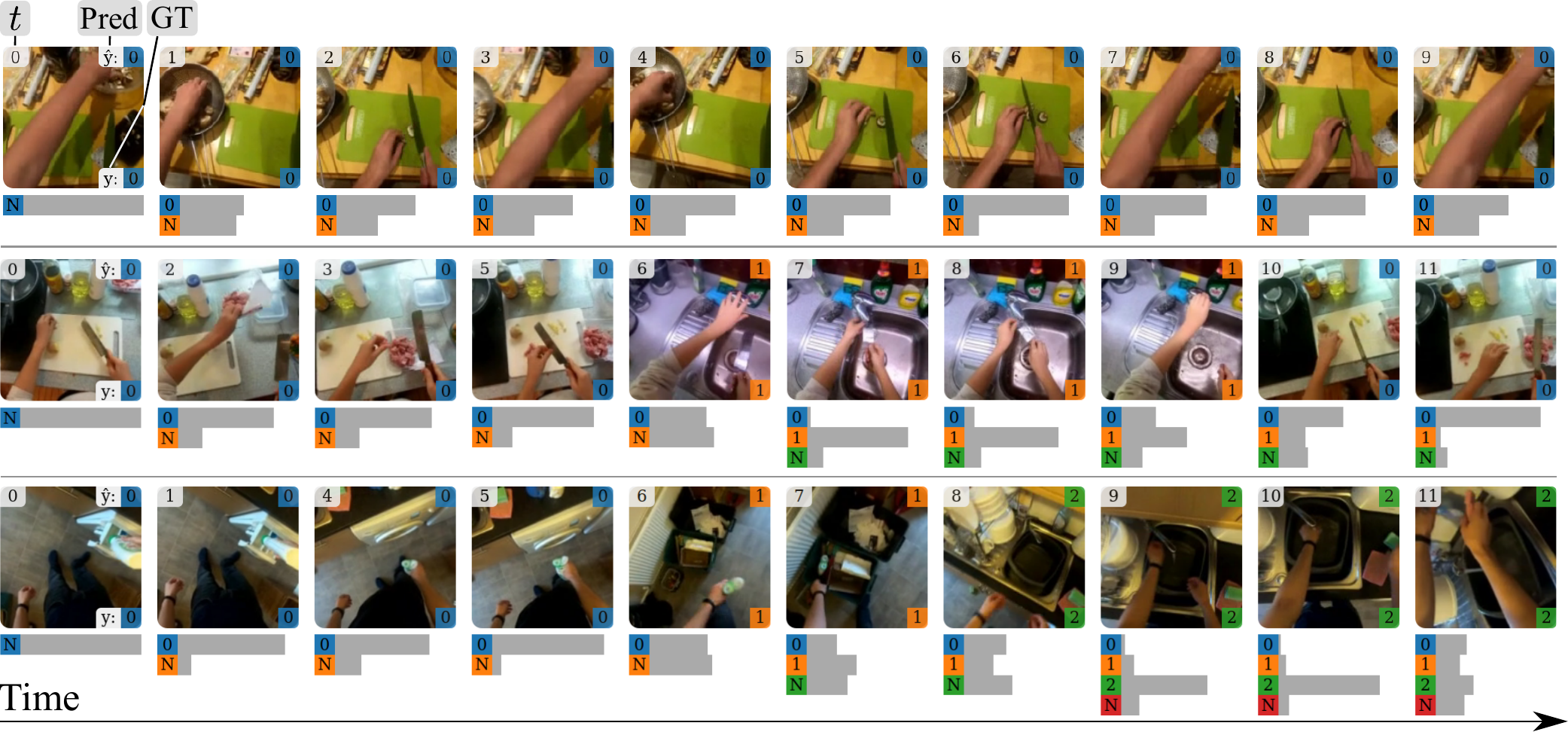}
  \vspace{-10pt}
  \caption{
    \textbf{Qualitative examples demonstrating UnweaveNet successfully unweaving 3 activity stories}.
    Decision probabilities $p_t$ are shown below each clip as a bar chart (N denotes a new thread).
    Top right corner indicates predicted thread, bottom right--ground-truth thread, and top left--clip index.
    \textbf{Top}~(1 thread story): \textcolor{unblue}{chopping mushrooms}.
    \textbf{Middle} (2 thread story): \textcolor{unblue}{dicing meat} (clips 0--5, 10--11) and \textcolor{unorange}{rinsing cleaver} (6--9).
    \textbf{Bottom} (3 thread story): \textcolor{unblue}{setting up washing machine} (0--5), \textcolor{unorange}{throwing bottle into recycling} (6--7) and \textcolor{ungreen}{washing hands} (8--11).
  }
  \label{fig:success-qualitative-examples}
  \vspace*{-12pt}
\end{figure}

\begin{figure}[t]
  \centering
  \includegraphics[width=\linewidth]{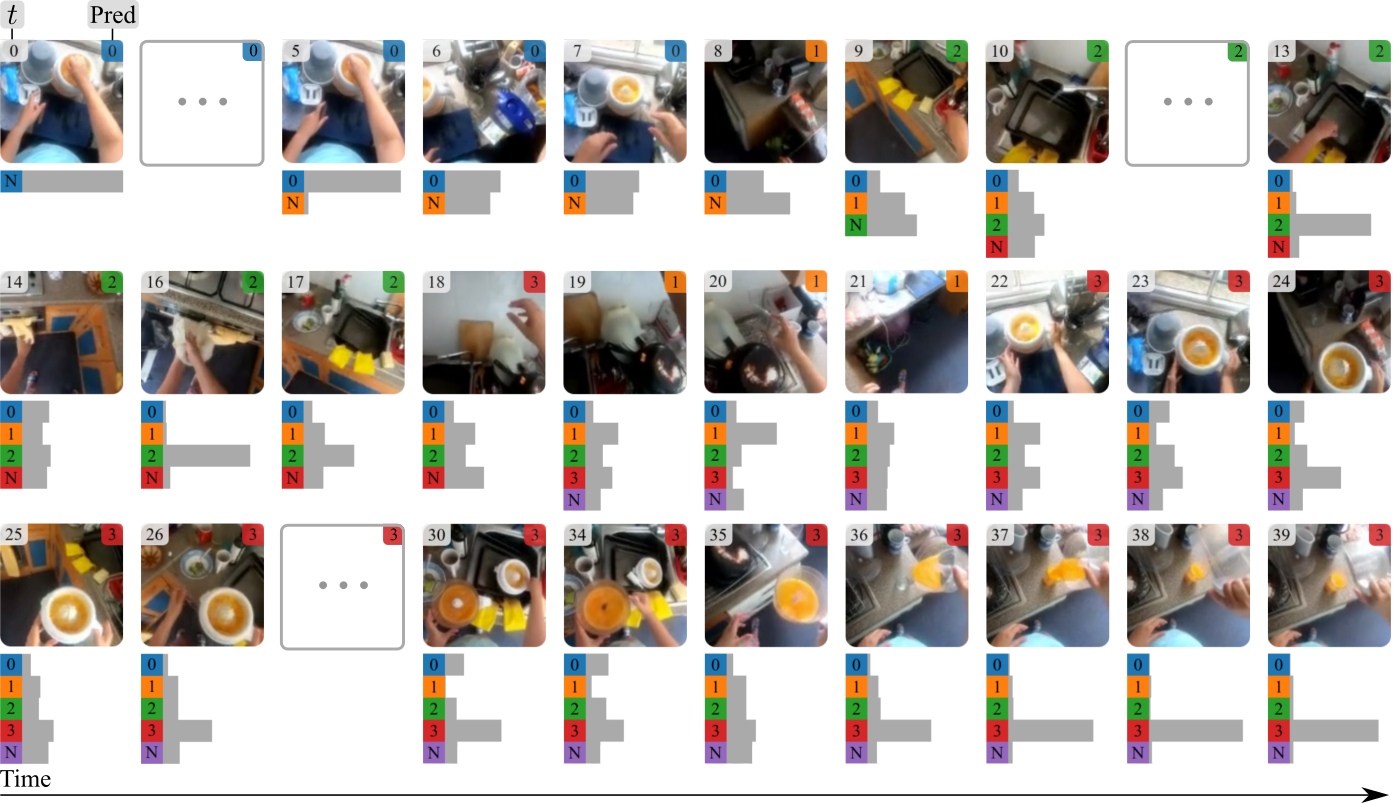}
  \caption{
    UnweaveNet represents this 40 clip sequence
    as  4 threads: \textcolor{unblue}{juicing the oranges} (0--7), \textcolor{ungreen}{washing hands} (9--17), \textcolor{unorange}{getting a glass} (19--21), and \textcolor{unred}{serving the orange juice} (22--39).
  }
  \label{fig:long-qualitative-example}
  \vspace*{-12pt}
\end{figure}

\begin{figure}[t]
  \includegraphics[width=\linewidth]{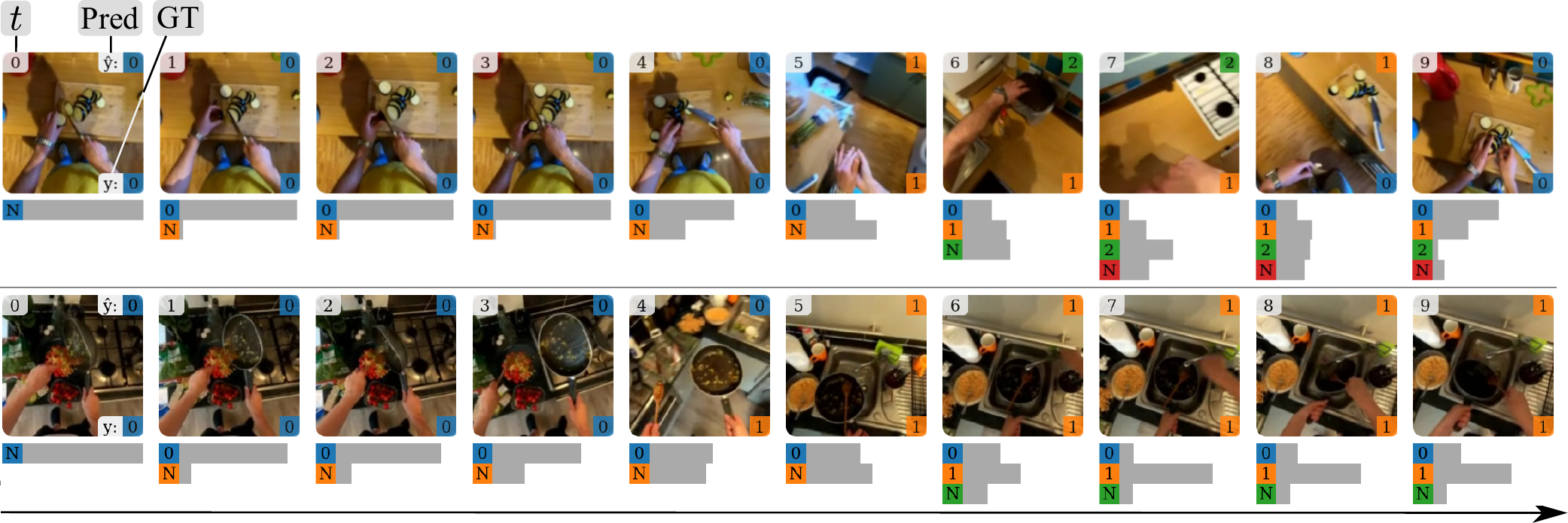}
  \caption[Qualitative examples demonstrating failure modes of UnweaveNet]{
    UnweaveNet's failure modes.
    \textbf{Top (over-segmentation)}: UnweaveNet separates the \textcolor{unblue}{chopping} activity (clips 0--4) from \textcolor{ungreen}{cleaning} (putting peelings into the bin) (clips 6--7) and correctly resumes the first thread (clip 9). However, an additional \textcolor{unorange}{incorrect thread} is created (clips 5 and 8)c to capture the transition.
    \textbf{Bottom (late-starts)}: two threads are recognised: \textcolor{unblue}{serving food} (clips 0--4) and \textcolor{unorange}{washing pan} (clips 5--9). However UnweaveNet leaves the serving thread one clip later than in the ground truth (clip 4).
  }
  \label{fig:qualitative-examples}
\end{figure}

Qualitative examples of the unweavings produced by UnweaveNet are provided in \cref{fig:success-qualitative-examples}, demonstrating the model's capability to unweave stories with varying numbers of threads.
\Cref{fig:long-qualitative-example} demonstrates how UnweaveNet is capable of unweaving a long sequence of 40 clips, demonstrating the ability of the model to create new threads of varying lengths.
\Cref{fig:qualitative-examples} demonstrates two failure modes of UnweaveNet: oversegmenting into too many threads and transitioning between threads slightly later than in the ground truth.
Further qualitative results can be found in \cref{appendix:additional-results}.

\begin{figure}
  \centering
  \includegraphics[width=\linewidth,clip,trim=0 50pt 0 0]{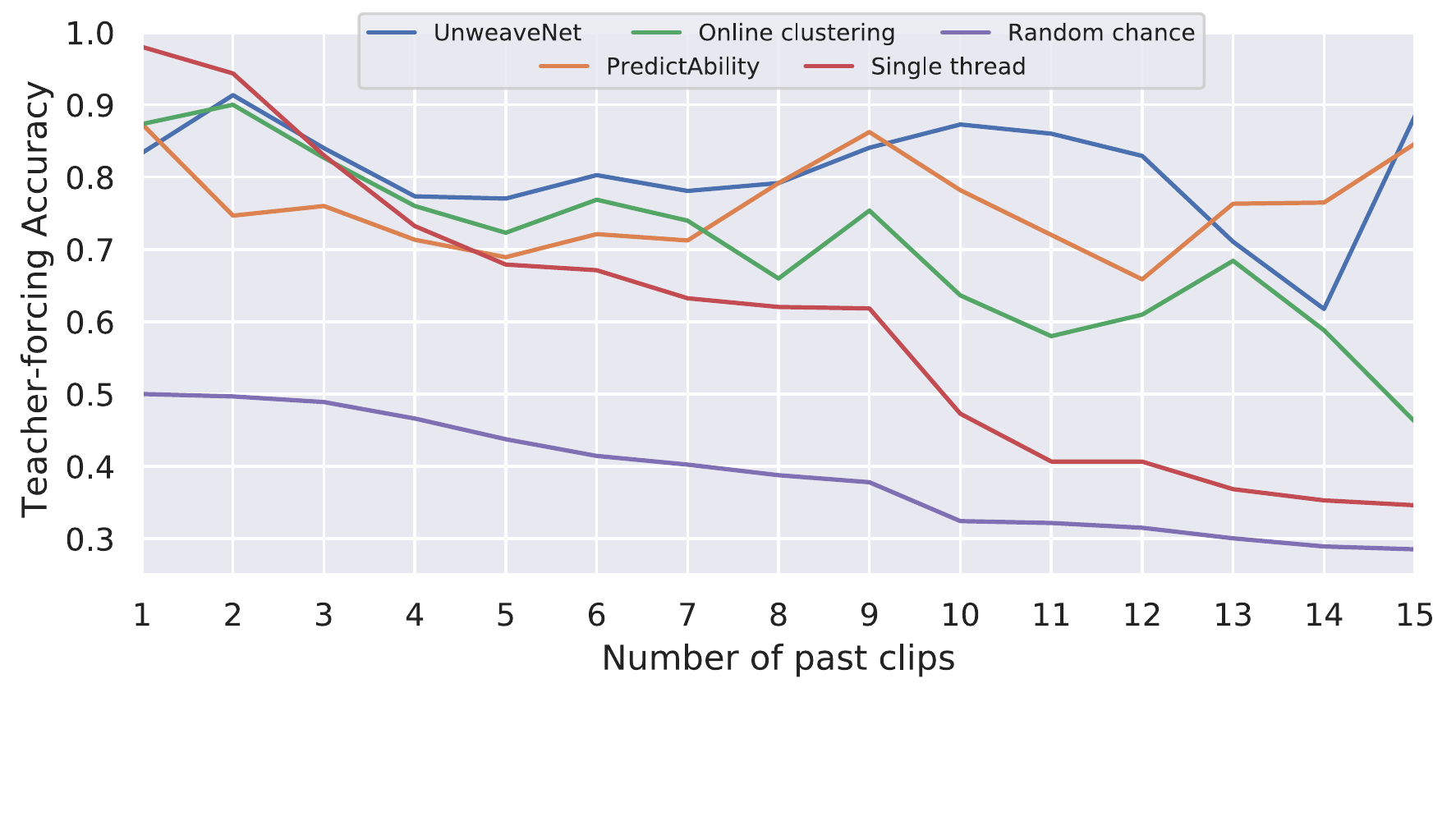}
  \vspace*{-10pt}
  \caption[Teacher-forcing accuracy of UnweaveNet compared to baselines]{Teacher-forcing accuracy by number of clips observed}
  \label{fig:online-performance}
\end{figure}

\Cref{fig:online-performance} shows how TFA varies during online predictions as more clips are observed. 
Initially, the single thread baseline has an easy task since few stories this short have more than a single thread, but from 4 clips onwards, UnweaveNet's performance gap over this baseline steadily increases, and the performance is robust as more clips are considered.
UnweaveNet outperforms the online clustering baseline from observing 4 clips onwards.
The teacher-forcing accuracy of the PredictAbility model is quite high due to two facts: the model cannot resume threads, and resuming a threads is a rarer event than continuing or starting a thread, therefore the model has fewer choices to take at each step and performs well on the more frequent scenarios.

\subsection{Ablation studies}
\label{chap:unweavenet:sec:ablation-studies}
Several ablation studies are conducted to determine the impact of the components of UnweaveNet on its behaviour. 
Each ablation study aims to answer a specific question.

\paragraph{How to best construct synthetic stories?}

Having established that pretraining on synthetic stories is beneficial (as was shown in \cref{tab:performance}), the best way of constructing them is investigated.
There are two hyperparameters to tune: the gap between clips within a synthetic thread and the number of synthetic threads forming the story.
As the gap between clips in a thread increases, the visual similarity between adjacent clips decreases, making the task of associating the clips together harder.
\Cref{tab:synthetic-story-ablation}a shows that increasing this gap up to 2 seconds is beneficial, but beyond this we observe a degradation.
Using a random gap between 2--4 seconds as an augmentation strategy was found to further boost performance over a fixed clip gap. 
This is the default configuration used throughout the remainder of the ablation studies.

When constructing synthetic stories, the number of threads is sampled uniformly from 1 to a maximum $N_\mathrm{max}$.
\Cref{tab:synthetic-story-ablation}b shows the RI increases as $N_\mathrm{max}$ is increased up to 4 threads, beyond which the performance decreases.
This drop can be attributed to the fact that threads are composed of fewer clips as the number of threads is increased in addition to the increased risk that some threads overlap in activity.

\begin{table}[t]
\centering
\begin{subtable}[t]{.45\linewidth}
\centering
    \begin{tabular}{@{}rr@{}}
    \toprule
    \makecell{Clip gap (s)}            & RI ($\uparrow$) \\
    \midrule
    0              & 60.6$\pm$0.6 \\
    1              & 73.1$\pm$0.3 \\
    2              & 74.6$\pm$0.7 \\
    4              & 73.7$\pm$0.7 \\
    2--4           & \textbf{75.1}$\pm 0.5$ \\ \bottomrule

    \end{tabular}
    \caption{Clip gap within thread.}
\end{subtable}
\begin{subtable}[t]{.45\linewidth}
  \centering
    \begin{tabular}{@{}rr@{}}
    \toprule
    \makecell{Max \# threads}  & RI ($\uparrow$) \\
    \midrule
    2                            & 69.0$\pm$0.6 \\
    3                            & 72.7$\pm$0.6 \\
    4                            & \textbf{75.1}$\pm$0.5 \\
    5                            & 74.0$\pm$0.8 \\
    6                            & 73.5$\pm$0.8 \\
    \bottomrule
    \end{tabular}
    \caption{Max \# threads.}
\end{subtable}
\caption{Synthetic story formation ablation study.}
    \label{tab:synthetic-story-ablation}
    \vspace*{-12pt}
\end{table}

\begin{table}
    \centering
    \begin{tabular}{@{}llrr@{}}
     \toprule
     $\selectmodel$   & $\updatemodel$ & RI ($\uparrow$) \\
     \midrule
     Linear Embedding & Last clip      & 73.7$\pm$0.7 \\
     Linear Embedding & GRU            & 74.3$\pm$0.3 \\
     Transformer      & Last clip      & 74.8$\pm$0.3 \\
     Transformer      & GRU            & \textbf{75.1}$\pm$0.5 \\


     \bottomrule
    \end{tabular}
    \caption{UnweaveNet architectural choices.}
    \vspace*{-12pt}
    \label{tab:model-ablation}
\end{table}
\paragraph{How to implement $\selectmodel$ and $\updatemodel$?}
The two versions of $\selectmodel$ introduced in \cref{chap:unweavenet:sec:method:controller}, $\linearselectmodel$ and $\transformerselectmodel$, are compared in  \cref{tab:model-ablation}.
The transformer based model proves superior to the linear embedding.
In a similar manner, two versions of $\updatemodel$ are compared using the recurrent update module based on a GRU vs.\ a linear projection of the last clip of each thread.
For the latter, the new clip representation overwrites the previous thread representation when performing an update.
The results demonstrate a small but consistent improvement when using the GRU update module.

\paragraph{}
An additional study on the effect of the weight in the loss function can be found in \cref{appendix:additional-results}.

\section{Conclusion}
\label{chap:unweavenet:sec:conclusion}
This paper introduced video \textit{unweaving}, the task of parsing a video online into its constituent activity \textit{threads}, accomplished by the introduction of a novel representation that models ongoing activity, operated by a neural controller, together called UnweaveNet.
UnweaveNet can handle resuming a thread when the video depicts a switch from one thread to a previously observed thread.
Moreover, it can be applied to variable-length videos, with memory requirements scaling linearly in the number of threads.
A dataset of activity stories was annotated and used to evaluate how UnweaveNet can be pretrained through self-supervision by sampling synthetic stories from untrimmed videos.

UnweaveNet has potential applications in assistive technologies as the activities are perceived online.
By focusing the experiments on egocentric footage, UnweaveNet is more suitable for sousveillance~\cite{mann2003_SousveillanceInventingUsing}, one's ability to monitor her/his activities, than surveillance, remote monitoring of others' activities.
However, in principle, the same approach can be adapted for monitoring other people's activity.

\vspace*{6pt}
\noindent \textbf{Acknowledgements.} This work used public dataset and was supported by EPSRC Doctoral Training Program, EPSRC UMPIRE~(EP/T004991/1), and NSF NRI Award~\#2132519.

\FloatBarrier
{\small \printbibliography}


\clearpage
\onecolumn
\appendix
\section*{Additional Material}

\noindent \textbf{Future Work.} In the paper, we make decisions online without recovering from errors. Recovery is a very interesting avenue for exploration, including soft assignment as in~\cite{graves2014_NeuralTuringMachines, grefenstette2015_Learningtransduceunbounded}, however these are harder to compare to prior work (e.g. Ego-Topo and Predictabilty from~\cite{nagarajan2020_EgoTopoEnvironmentAffordances}). An exploration of the right metrics and baselines is needed to consider recovery. 

These appendices provide additional material on UnweaveNet.
\Cref{appendix:synthetic-stories} explains how synthetic stories are constructed, detailing the exact procedure used to sample them.
\Cref{appendix:activity-stories} covers the procedure used to annotate activity stories, demonstrating the user interface that was developed for this.
\Cref{appendix:metrics} explains the metrics used for evaluating the unweaving task.
\Cref{appendix:additional-results} presents additional and extended experimental results of UnweaveNet.

\section{Constructing synthetic stories}
\label{appendix:synthetic-stories}
UnweaveNet is pretrained in a self-supervised manner by learning to unweave synthetic stories constructed via a randomised sampling procedure applied to long video.
These synthetic stories aim to pose a similar challenge to unweave as activity stories, albeit in a somewhat label-noisy manner since they are constructed through a fully automated process.

Synthetic stories are composed of a number of randomly sampled subsequences of different lengths, termed synthetic threads, that are randomly woven together.
A graphical overview of this process is given in \cref{fig:synthetic-story-formation}.
Synthetic threads are sampled from the same video as sampling them from different videos would result in a story that is trivial to unweave due to the large visual differences between videos.
Threads are sampled such that they are at least a minimum distance away from one another and are assumed to depict distinct activities.

\paragraph{Sampling synthetic threads}
Building a synthetic story starts by obtaining a number of synthetic threads to weave together.
These are obtained by sampling a number of sequences of clips of varying lengths from distinct non-overlapping locations within a video.
First, the number of clips $T$ that will comprise the story is decided.\footnote{$T$ is fixed across stories to facilitate batch-based training.}
Then, the number of threads $n$ in the story is sampled from the uniform distribution $\mathcal{U}\{1, N_{\mathrm{max}}\}$ where $N_\mathrm{max}$ is a specified upper bound.
Next, the number of clips $m_i$ comprising thread $i$ is determined by uniformly sampling $n$ non-zero positive integers $\left(m_i \right)^{n}_{i=1}$ such that $\sum_{i=1}^n m_i = T$ using the method of \textcite{smith2004_Samplinguniformlyunit}.
The starting location of the threads within the video are then sampled randomly from a uniform distribution with the constraint that the threads are at least a minimum distance away from one another.
The clips $\mathcal{V}^i = \left(v^i_t\right)_{t=1}^{m_i}$ comprising thread $i$ are then sampled starting from the thread's starting location, obtained in the previous step.
Adjacent clips within a thread are separated by a small random gap sampled from $\mathcal{U}\{G_{\mathrm{min}}, G_{\mathrm{max}}\}$  where $G_\mathrm{min}$ and $G_\mathrm{max}$ denote the specified minimum and maximum gap, further increasing the difficulty of unweaving the resulting synthetic stories.

\paragraph{Weaving threads}
Once the threads have been obtained, they need to be woven together to form a synthetic story.
This is performed such that the within-thread temporal-ordering of clips is not disrupted.
In other words, given a thread $i$ composed of the clips $\left(v^i_1, v^i_2\right)$, $v^i_1$ will always appear before $v^i_2$ in the synthetic story, and so on for threads with more clips.
Weaving is accomplished by building a template for how the clips will be ordered relative to one-another, based on the number of clips $m_i$ comprising each thread.
First, a vector $q\in \{1..n\}^T$ of repeated thread indices is constructed, where there are $m_i$ repeats of each thread index $i$:
\begin{equation}
q = (\underbrace{1, \ldots, 1}_{\text{$m_1$ times}},\; \underbrace{2, \ldots, 2}_{\text{$m_2$ times}},\; \ldots,\; \underbrace{n, \ldots, n}_{\text{$m_n$ times}}).
\end{equation}
Next, a random permutation of $q$ is taken, denoted $\tilde{q}$, which forms the template used to weave the threads' clips together into the story $v$.
Each clip $v_j$ in the story can thus be defined as
\begin{equation}
  v_j = v^{\tilde{q}_j}_{o_j},\quad o_j = \sum_{k=1}^j \mathbb{I}\left[ \tilde{q}_j = \tilde{q}_k \right].
\end{equation}
where $o_j$ represents the number of clips in the story from thread $\tilde{q}_j$ up to time $j$.
The following example demonstrates this procedure to aid comprehension.
Let the number of total number of clips in the story be $T = 10$, the number of threads be $n = 4$ and the number of clips within each thread be $m = (3, 5, 4)$.
The duplicated thread indices vector $q$ is formed:
\begin{equation}
q = (1, 1, 1, 2, 2, 2, 2, 2, 3, 3, 3, 3).
\end{equation}
and is randomly permuted to obtain
\begin{equation}
\tilde{q} = (1, 2, 1, 2, 2, 3, 3, 2, 2, 3, 3, 1),
\end{equation}
which is then used to build the story
\begin{equation}
    v = \left(v^1_1,  v^2_1,  v^1_2,  v^2_2,  v^2_3,  v^3_1,  v^3_2,  v^2_4,  v^2_5,  v^3_3,  v^3_4,  v^1_3\right).
\end{equation}
Note the within-thread temporal-ordering of clips has not been violated in $v$.

\paragraph{Error analysis in synthetic stories}
Studying Table~\ref{tab:activity-dataset-stats} where thread statistics are given, two randomly sampled consecutive clips belong to the same thread 95.2\% of the time.
Accordingly, we estimate that $\sim$95\% of all \textcolor{ungray}{C} decisions in synthetic stories to be correct.
Also, we note that synthetic threads are sampled from distinct locations in the same video 
to increase the realism. We enforce 60s gap and accordingly estimate 99\% of all \textcolor{ungreen}{N} decisions to be also correct. 

\section{Annotating activity stories}
\label{appendix:activity-stories}

\begin{figure}[t]
  \begin{tikzpicture}
    \node[anchor=south west,inner sep=0](image) at (0, 0){\includegraphics[width=\linewidth]{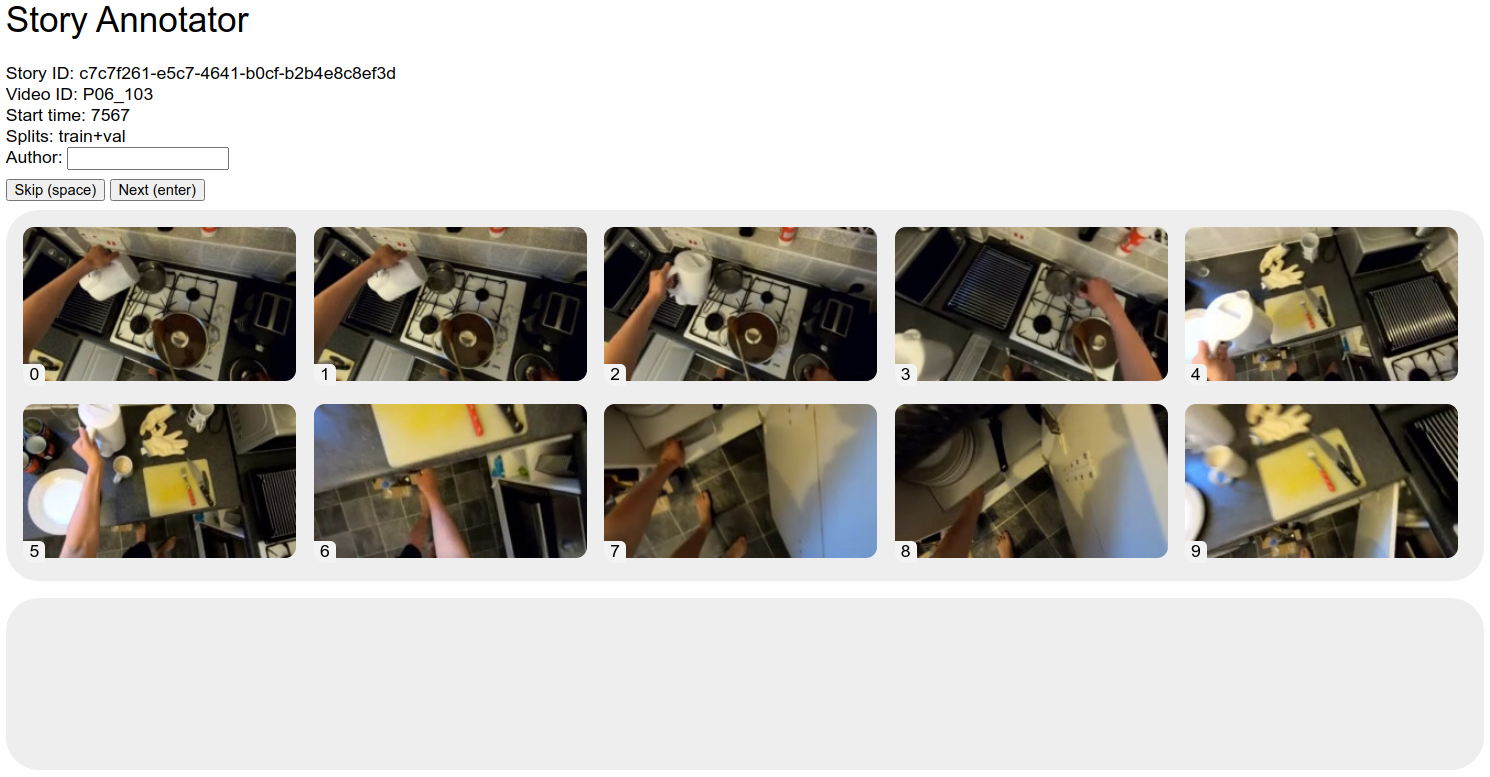}};
    \begin{scope}[x={(image.south east)},y={(image.north west)}]
      \draw[->,gray,thick] (0.3, 0.285) to[out=-120, in=45] (0.05,0.12);
      \draw[->,gray,thick] (0.5, 0.285) to[out=-120, in=45] (0.25,0.12);
      \draw[->,gray,thick] (0.7, 0.285) to[out=-120, in=45] (0.45,0.12);
      \draw[->,gray,thick] (0.9, 0.285) to[out=-120, in=45] (0.65,0.12);
    \end{scope}
  \end{tikzpicture}
    \caption[Story Unweaving annotation tool (pre-interaction)]{
      \textbf{Story unweaving annotation tool (pre-interaction):}
       Initial state of the UI when presented with a new video.
      \tikz[baseline=(current bounding box.base)] \draw[->,gray,thick] (0ex, 3pt) to (10pt, 3pt); denotes a user dragging a clip into a new thread track (the empty grey rectangle).
    }
    \label{fig:story-unweaver-ui-before}
\end{figure}
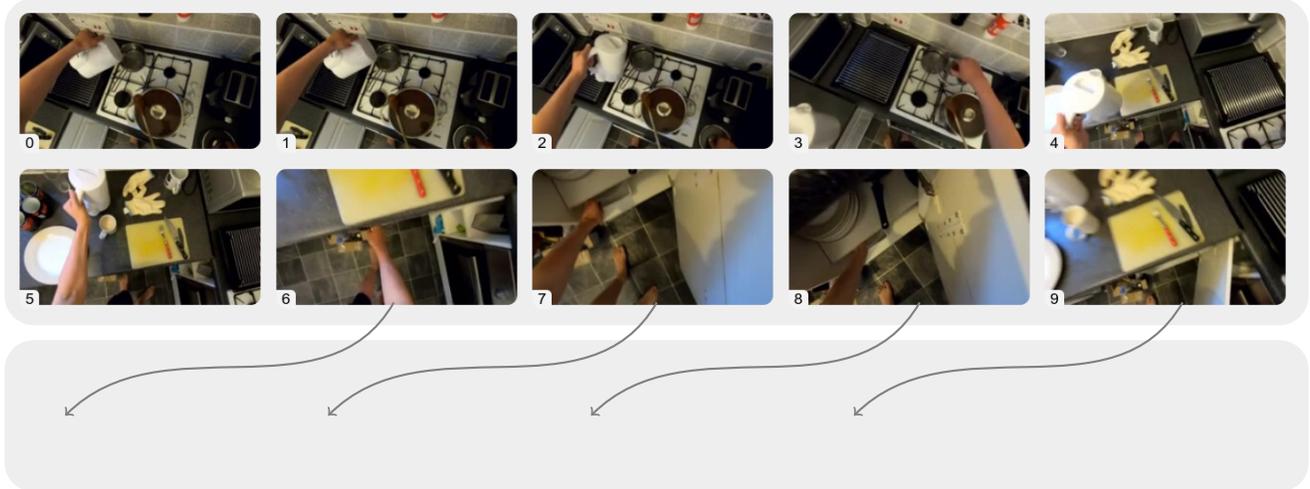

\begin{figure}[t]
\includegraphics[width=\linewidth]{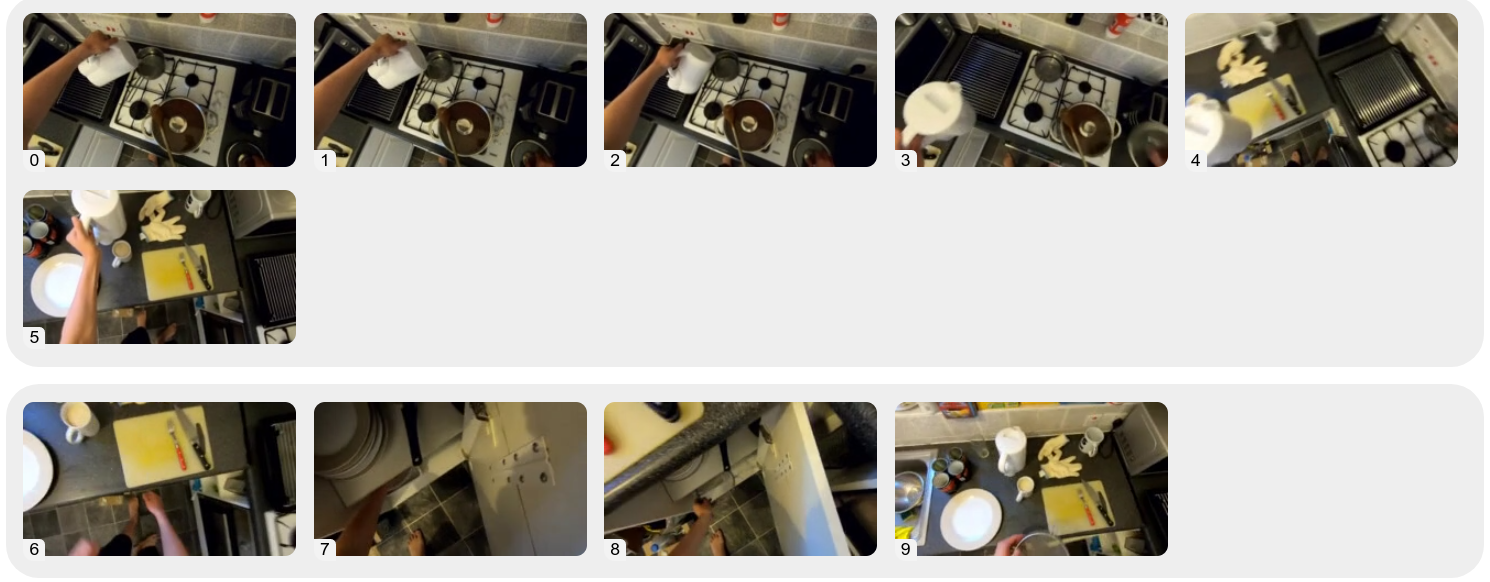}
\caption[Story Unweaving annotation tool (post-interaction)]{
  \textbf{Story unweaving annotation tool (post-interaction).}
  State of UI after the annotator has dragged the clips into a new thread (interaction depicted in \cref{fig:story-unweaver-ui-before}).
  The ordering of the clips is always preserved; \ie{} the tool prevents users from reordering the clips so that one that came at time $t$ in the source video comes before one at $t-1$.
}
\label{fig:story-unweaver-ui-after}
\end{figure}

A web-based unweaving tool (shown in \cref{fig:story-unweaver-ui-before,fig:story-unweaver-ui-after}) was developed to enable the annotation of stories within video.
The tool randomly samples a sequence of video from a collection of videos and tasks the annotator with manually unweaving the sequence.
The sampled sequence of video is broken up into a fixed number of clips which are displayed within a single track that represents a thread (grey box  in \cref{fig:story-unweaver-ui-before}).
The annotator can then drag and drop clips (interaction denoted by the dark grey arrows in \cref{fig:story-unweaver-ui-before}) into a new track to form a new thread (the result of which is shown in \cref{fig:story-unweaver-ui-after}).
Once complete, the tool saves the annotation to a database and ensures that this portion of video is not shown again to another annotator.
Annotators were provided with the following instructions to use the tool:
\begin{displayquote}
\emph{Thread annotation:}
We have sampled consecutive clips from a video (labelled 0--9) and the interface allows grouping these into activity `threads' (sequence of clips sharing a common goal).
Your task is to identify occurrences where the person is changing the course-of-action or goal (\eg{} switch from washing-up to cooking, or from preparing-food to returning-items-to-cupboards).
Once identified, the interface allows you to group all clips belonging to one goal in the same thread by dragging other clips to a new thread.
If the person returns to an earlier goal (\eg{} Goal switches $A \rightarrow B \rightarrow A$), make sure to keep all $A$ clips together in one thread.

\emph{Using the annotator:}
Your task is to drag video clips onto a new track (denoted by the gray background) when you detect a change in the course-of-actions.
If the example is ambiguous or you cannot understand the activity from the clips or the videos are too dark to see, click skip (alternatively press \texttt{<space>}).
Click next (alternatively press \texttt{<enter>}) to save and continue on to the next example.
\end{displayquote}

\section{Experimental details}
\label{appendix:experimental-details}
This section describes the experimental set up use to train and evaluate UnweaveNet and the baselines.

\paragraph{Data}
Videos are encoded to 16FPS, resized to a height of 112px, and center-cropped.
Synthetic stories are constructed from 1s long clips which are separated from adjacent clips in a synthetic thread by a gap of between 2--4s.
To lower the chance of sampling threads sharing the same activity, threads are separated by at least 60s.
Each synthetic story is composed of 10 clips and is produced by weaving between 1--4 synthetic threads, unless stated otherwise.
Synthetic stories are sampled randomly from the training videos of EPIC-KITCHENS, thus the model is trained on a practically infinite number of synthetic stories.
Videos shorter than 3\textonehalf{} mins long are discarded, as they are insufficiently long to sample synthetic stories from.
In all experiments using synthetic story pretraining, around 800K synthetic stories are sampled (8M clips, 50K batches, each containing 16 stories).

\paragraph{UnweaveNet}
The backbone network used to extract clip features is a top-heavy 3D ResNet-18 pretrained on Kinetics~\cite{kay2017_KineticsHumanAction} using the DPC self-supervised objective~\cite{han2019_VideoRepresentationLearning}.
The backbone's features are average pooled both spatially and temporally to produce a single vector per clip.
Clip and thread features have dimensions $C = D = 256$ respectively and are embedded into to an $E = 256$ dimensional space for $\linearselectmodel$ or $E = 512$ for $\transformerselectmodel$.
The thread representation update module $\updatemodel$ is instantiated using a single layer GRU~\cite{cho2014_PropertiesNeuralMachine} with a hidden/output dimension of 256.
The transformer-based controller $\transformerselectmodel$ uses 1 layer, 4 heads, a model dimension of 512, and a feed forward MLP dimension of 2048 with a dropout of 0.2 (applied only during pretraining).
The new thread token \texttt{[NT]} is defined as a learnt latent vector, and the empty thread representation as $z^* = \mathbf{0}$  (using a learnt latent vector was also experimented with, but it did not yield any improvements).
The softmax temperature $\tau$ is set to $0.05$.

UnweaveNet is pretrained on synthetic stories, generated on the fly, for 50k steps (determined by measuring performance on a validation set of synthetic stories).
The learning rate is set to $2\times 10^{-4}$ for 25k steps and then dropped to $2\times 10^{-5}$ for the remaining 25k steps.
Finetuning is conducted using the learning rate $2\times 10^{-5}$ and proceeds for up to 1k steps, with early stopping based on validation split performance.
Adam~\cite{kingma2015_AdamMethodStochastic} is used to optimise the models with a batch-size of 16 stories on 2x NVIDIA RTX-2080 Tis.
Each training step takes around 1.3s; therefore pretraining takes around 18 hours and finetuning over 5 random seeds just under 2 hours.
Results are reported as the mean over 5 different seeds (consistent across experiments) used in finetuning, starting from a single synthetic story pretrained checkpoint.
Horizontal flipping is used for data augmentation, applied consistently to all clips within a story.
Other augmentation strategies, including random crops and color distortion, were attempted, but didn't improve performance.

\paragraph{Baselines}
Both PredictAbility~\cite{shou2021_GenericEventBoundary} and the online clustering baseline use features extracted from a top-heavy 3D ResNet18 trained with the DPC objective on Kinetics, the same as used for the clip backbone in UnweaveNet.
Hyperparameters for the following baselines are chosen by performing a hyperparameter search optimising for the average RI across the validation set.
PredictAbility uses a temporal stride of 2, a window of 10 frames each side of the candidate boundary, and a $\sigma$ of 15 frames for the Laplacian of Gaussian kernel.
The online clustering baseline uses a similarity threshold of 0.645 above which the clip is judged a continuation of the thread.
EGO-TOPO uses a window size of 8 frames, a lower threshold of 0.4, and an upper threshold of 0.6.
The localisation network used is the same as the one released by the authors.\footnote{EGO-TOPO source code and models: \url{https://github.com/facebookresearch/ego-topo}}

\section{Metrics}
\label{appendix:metrics}


\paragraph{Rand Index (RI)}
\label{appendix:metrics}
As unweaving is a form of clustering, a clustering metric is used as the main assessor of the quality of the unwoven threads.
The Rand index~\cite{rand1971_ObjectiveCriteriaEvaluation}, a frequently used metric to assess the similarity of one clustering to another, fulfils this role.
It computes the percentage of correct pair-wise decisions.

The Rand index is best understood by examining the meaning of true/false positives/negatives in this setting.
Given a story that is annotated with a ground-truth set of threads $G$ and that has been unwoven by a model into a set of threads $P$, both defined as partitions over the set of clips comprising the story, the definition of true/false positives/negatives are:
\begin{itemize}
    \item True positives $\tp(G, P)$: the number of pairs of clips that are in the same thread in $G$ and in the same thread in $P$.
    \item False positives $\fp(G, P)$: the number of pairs of clips that are in different threads in $G$ but in the same thread in $P$.
    \item True negatives $\tn(G, P)$: the number of pairs of clips that are in different threads in $G$ and also in different threads in $P$.
    \item False negatives $\fn(G, P)$: the number of pairs of clips that are in the same thread in $G$ but in different threads in $P$.
\end{itemize}
These can then be used to define the Rand index:
\begin{equation}
  \label{eq:rand-index}
  \operatorname{RI}(G, P) = \frac{\tp(G, P) + \tn(G, P)}{\tp(G, P) + \fp(G, P) + \tn(G, P) + \fn(G, P)}.
\end{equation}
Note that the denominator in \cref{eq:rand-index} is equal to ${T \choose 2}$ where $T$ is the total number of clips.
The Rand index is computed for each story and is averaged over all stories in the test set to produce a single score.

The expected Rand index $\expectation{P \sim R(T)}{\operatorname{RI}(G, P)}$ for the na\"ive baselines can be computed by defining their corresponding distribution $R(T)$ over partitions of $T$ clips.
\textcite{gates2017_impactrandommodels} provide closed form expressions for both na\"ive baselines.
For the single-thread baseline $R^1$,
\begin{equation}
  \expectation{P\sim R^1(T)}{\operatorname{RI}(G, P)} =
  \frac{\sum_i {|\mathcal{V}^i|  \choose 2}}{{T \choose 2}}
\end{equation}
where $|\mathcal{V}^i|$ the number of clips in the $i$-th thread of partition $P$.
For the random-chance baseline $R^\mathrm{all}$,
\begin{equation}
  \expectation{P\sim R^\mathrm{all}(T)}{\operatorname{RI}(G, P)} =
  \frac{B(T - 1)}{B(T)}\frac{\sum_i {|\mathcal{V}^i| \choose 2}}{{T \choose 2}} +
  \left( 1 - \frac{B(T - 1)}{B(T)} \right) \left( 1 - \frac{\sum_i{|\mathcal{V}^i| \choose 2}}{{T \choose 2}} \right),
\end{equation}
where $B(T)$ is the $T$-th Bell number, which counts how many possible ways there are of partitioning a set of $T$ objects into non-empty subsets.


\paragraph{Teacher-forcing accuracy (TFA)}
The teacher-forcing accuracy measures the proportion of clip decisions that were made correctly, assuming that all the past decisions up to that point were correct.
The teacher-forcing accuracy is broken down by $n$, the number of threads that have been observed up to and including time $t$.

To describe how teacher-forcing accuracy is computed, it is necessary to define the set of story prefixes $\mathcal{P}_n$, story subsequences starting from the first clip onwards that contain exactly $n$ threads.
These are defined as
\begin{equation}
  \mathcal{P}_n = \left\{ v_{1:t} \,|\, v \in \mathcal{X} \land N(v_{1:t}) = n \land 1 \leq t \leq T(v) \right\},
\end{equation}
where $\mathcal{X}$ represents the dataset of stories, $N(v_{1:t})$ denotes the number of threads in the ground truth for $v$ up to and including time $t$, and $T(v)$ denotes the number of clips in $v$.
The teacher-forcing accuracy for clip decisions where exactly $n$ threads have been observed can then be defined as
\begin{equation}
  \label{eq:oa-model}
  \operatorname{TFA}(\mathcal{P}_n) = \frac{1}{|\mathcal{P}_n|}\sum_{v \in \mathcal{P}_n} \mathbb{I}\left[ \hat y'_{T(v)}(v) = y_{T(v)}(v) \right] \mathbb{I}\left[ |v| > 1\right],
\end{equation}
where $\hat y'_t(v)$ denotes the decision produced by the model at time $t$ on $v$ when it is run using teacher-forcing and $y_{t}(v)$ denotes the ground-truth thread-index at time $t$ in story $v$.
The $|v| > 1$ condition in \cref{eq:oa-model} removes stories composed of a single clip where all methods will trivially make the correct decision.
For UnweaveNet, the thread bank is populated using $\updatemodel$ according to the ground-truth thread assignments $y_{1:T(v)-1}$ to produce $z_{T(v)}$.
The controller $\selectmodel$ is then fed $v_{T(v)}$ and $z_{T(v)}$ to determine $\hat{y}'_{T(v)}(v)$.

An analogous approach is taken for the online clustering baseline, populating the clusters according to the ground-truth.
PredictAbility's decision is judged to be correct at each time step if the ground truth has a thread continuation and the model does not detect a transition, or if there is a new thread in the ground truth and the model detects a transition.
It is not possible to evaluate the TFA of the EGO-TOPO model with the provided implementation.

To compute the teacher-forcing accuracy for the na\"ive baselines, the term $\mathbb{I}[\hat{y}'_{T(v)}(v) = y_{T(v)}(v)]$ in \cref{eq:oa-model} is replaced with the probability $p^R_t(v)$  of selecting the correct thread at time $t$ in video $v$ for a baseline model $R$.
For the single thread baseline,
\begin{equation}
  \label{eq:oa-single-thread}
  p^1_t(v) =
  \begin{cases}
    1/N(v_{1:t-1}) & y_t \mathrm{\ is\ an\ existing\ thread}\\
    0 & y_t \mathrm{\ is\ a\ new\ thread}.
  \end{cases}
\end{equation}
The reasoning behind the first case in \cref{eq:oa-single-thread} is that there may be more than one thread in the ground-truth unweaving up to time $t$.
Because of this, the baseline has to make a choice which thread the clip will join, so the choice is made randomly.
However, this baseline is unable to start a new thread (case two).

The formulation for the random chance baseline is simpler, since the probability that the correct decision is made is uniform across the possible options: $p^\mathrm{all}_t = 1/(N(v_{1:t-1}) + 1)$.
Note that the denominator is $N(v_{1:t-1}) + 1$ and not $N(v_{1:t-1})$ as there is the additional option of starting a new thread at each time-step.

\paragraph{Difference in number of threads ($\Delta N$)}
It is informative to know whether the model oversegments, creating more threads than in the ground truth, or undersegments, creating fewer threads than in the ground truth.
Computing the difference $\Delta N = \hat{N} - N$ between the number of threads $\hat{N}$ detected by a model and the number of threads $N$ in the ground truth reveals this.

For the single-thread baseline, the number of predicted threads is always set to one.
For the random-chance baseline, a closed form expression can be derived using Stirling numbers of the second kind which compute the number of ways to partition $n$ items into $k$ non-empty subsets denoted $S(n, k)$.
The number of threads this baseline produces over a video with $T$ clips is simply a weighted average over partitions of size $n$:
\begin{equation}
\hat{N}^\mathrm{all} = \frac{1}{\sum_{n=1}^{T}S(T, n)}\sum_{n=1}^{T} S(T, n)n.
\end{equation}
This metric is computed for each story and averaged across all stories to produce a single score across the dataset.

\section{Additional results}
\label{appendix:additional-results}


\paragraph{Additional qualitative results}
\begin{figure}[ht]
  \centering
  \textbf{Single thread}
  \vskip5pt
  \includegraphics[width=\linewidth]{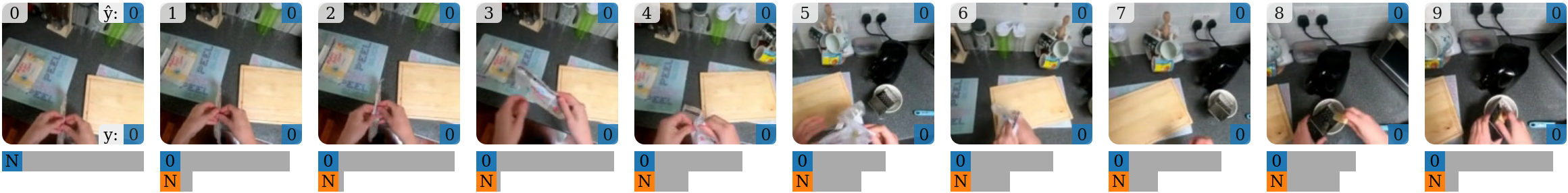}
  \includegraphics[width=\linewidth]{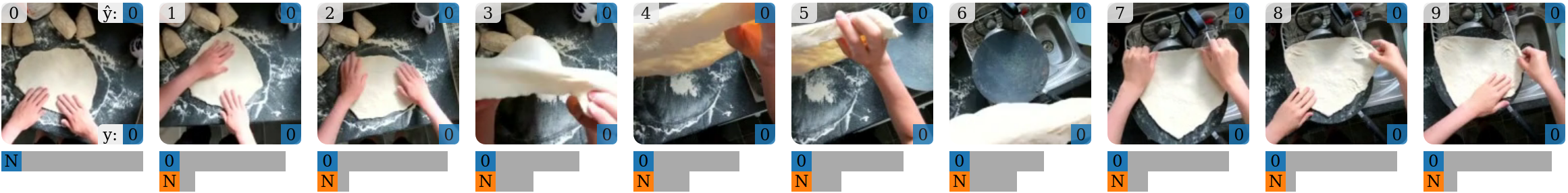}
  \includegraphics[width=\linewidth]{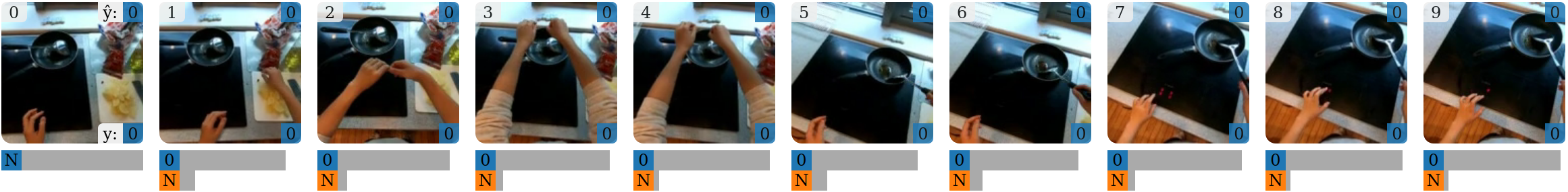}
  \vskip5pt
  \hrule
  \vskip5pt
  \textbf{Two threads}
  \vskip5pt
  \includegraphics[width=\linewidth]{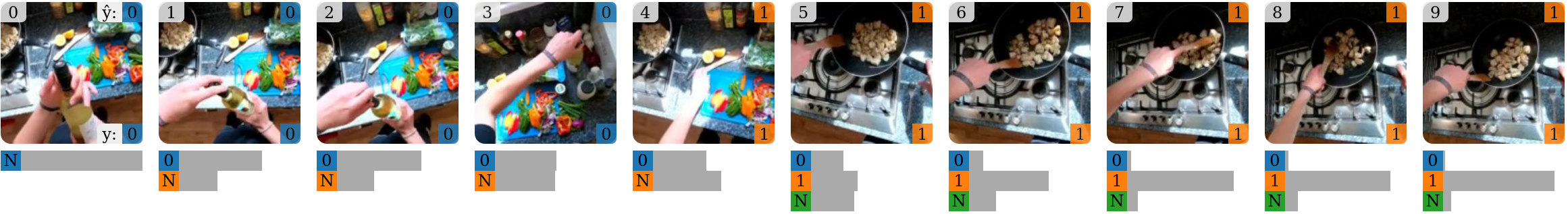}
  \includegraphics[width=\linewidth]{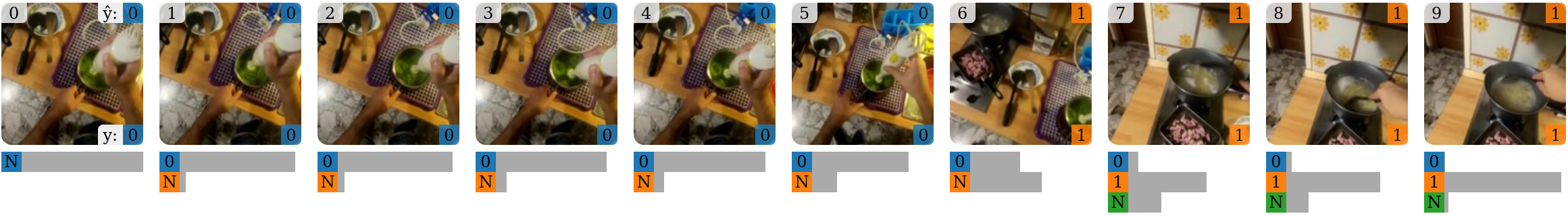}
  \vskip5pt
  \hrule
  \vskip5pt
  \textbf{Three threads}
  \vskip5pt
  \includegraphics[width=\linewidth]{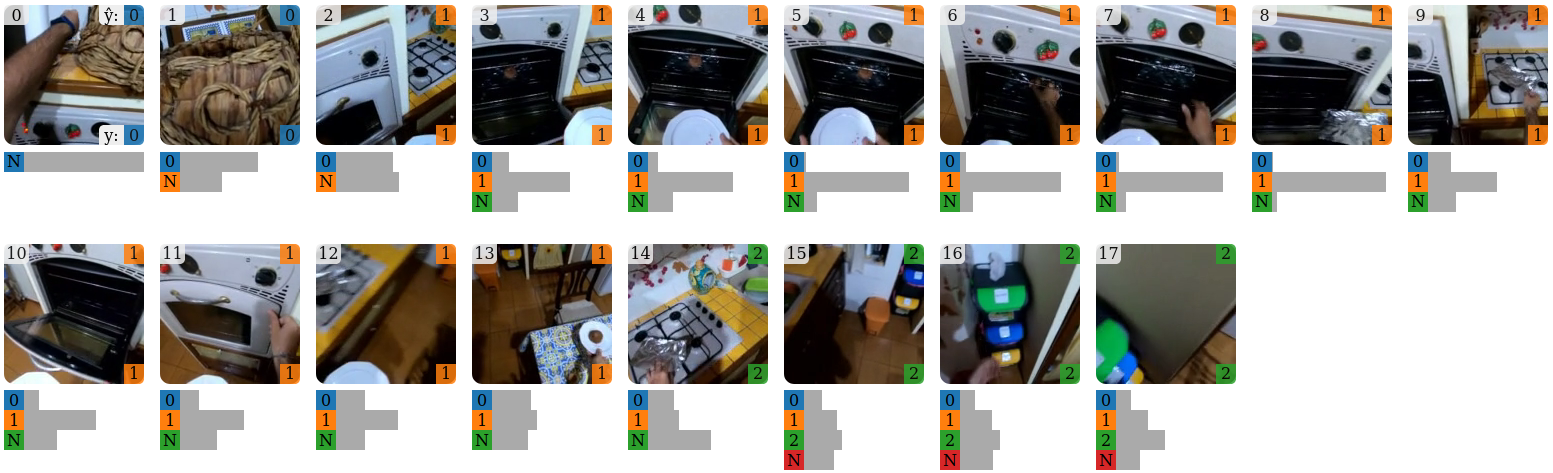}
  \caption{
    Additional examples demonstrating UnweaveNet successfully unweaving videos.
    See \cref{fig:success-qualitative-examples} for a legend.
    \textbf{Single thread (top)}
    row one: \textcolor{unblue}{preparing grated cheese};
    row two: \textcolor{unblue}{preparing pizza dough};
    row three: \textcolor{unblue}{stir frying herbs}.
    \textbf{Two threads (middle)}:
    row one: \textcolor{unblue}{closing and putting away bottle}, \textcolor{unorange}{stirring contents of pan};
    row two: \textcolor{unblue}{blending soup}, \textcolor{unorange}{stirring pasta}.
    \textbf{Three threads (bottom)}: \textcolor{unblue}{turning oven off}, \textcolor{unorange}{serving baked potato}, \textcolor{ungreen}{recycling used tin foil}.
  }
  \label{fig:appx:additional-success-cases}
\end{figure}

\begin{figure}[ht]
  \centering
  \includegraphics[width=\linewidth]{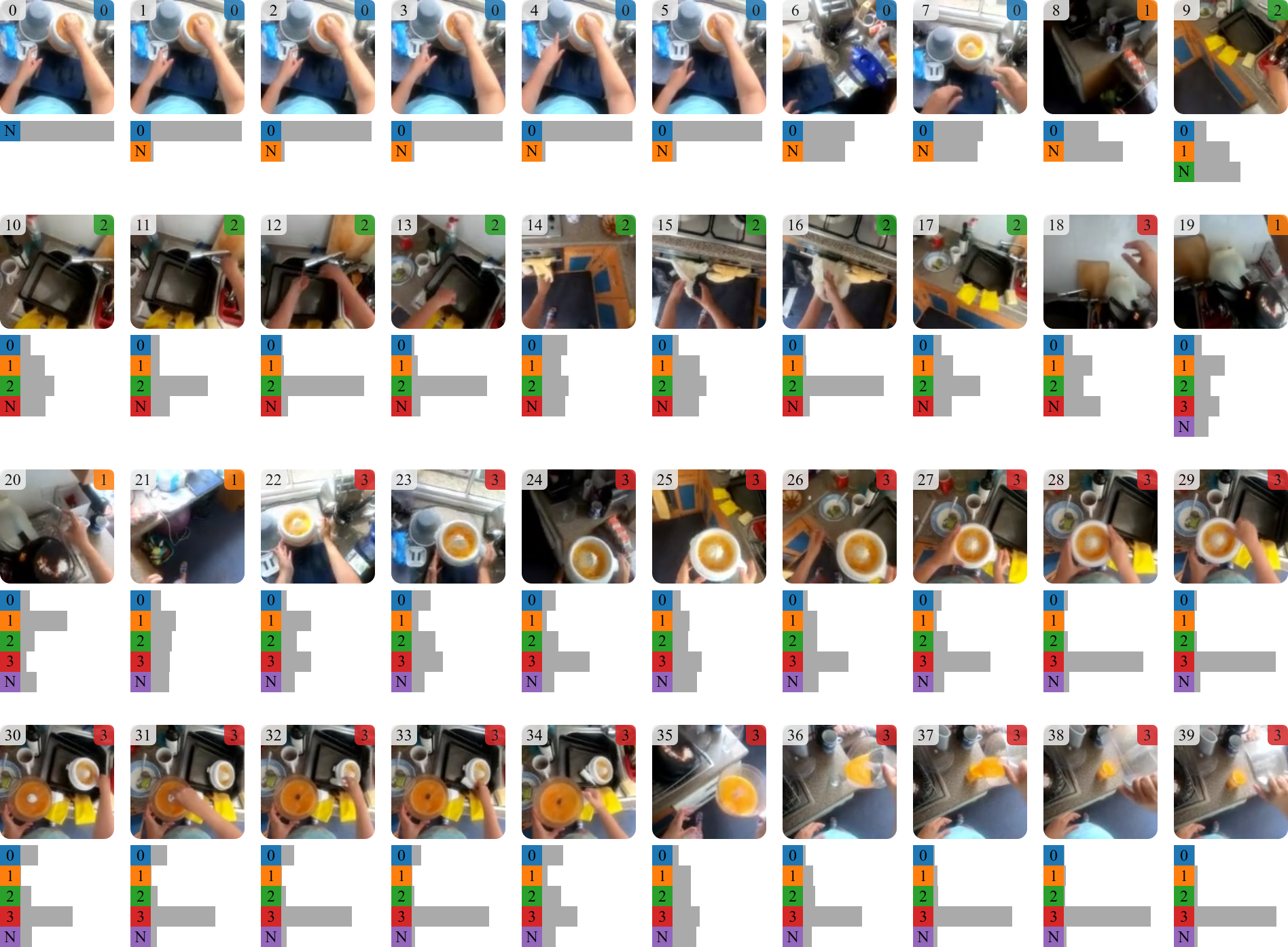}
  \caption{Untruncated version of \cref{fig:long-qualitative-example}.
    UnweaveNet represents this video as  4 threads: \textcolor{blue}{juicing the oranges} (0--7), \textcolor{green}{washing hands} (9--17), \textcolor{orange}{getting a glass} (19--21), and \textcolor{red}{removing the juicer and serving the orange juice} (22--39).
  }
  \label{fig:appx:long-qualitative-example-full}
\end{figure}

We present an additional selection of successful predictions by UnweaveNet~\cref{fig:appx:additional-success-cases}.
We provide the full, non-truncated, version of \cref{fig:long-qualitative-example} in \cref{fig:appx:long-qualitative-example-full}.

\begin{figure}[t]
\begin{minipage}{0.49\linewidth}
  \includegraphics[width=1\linewidth]{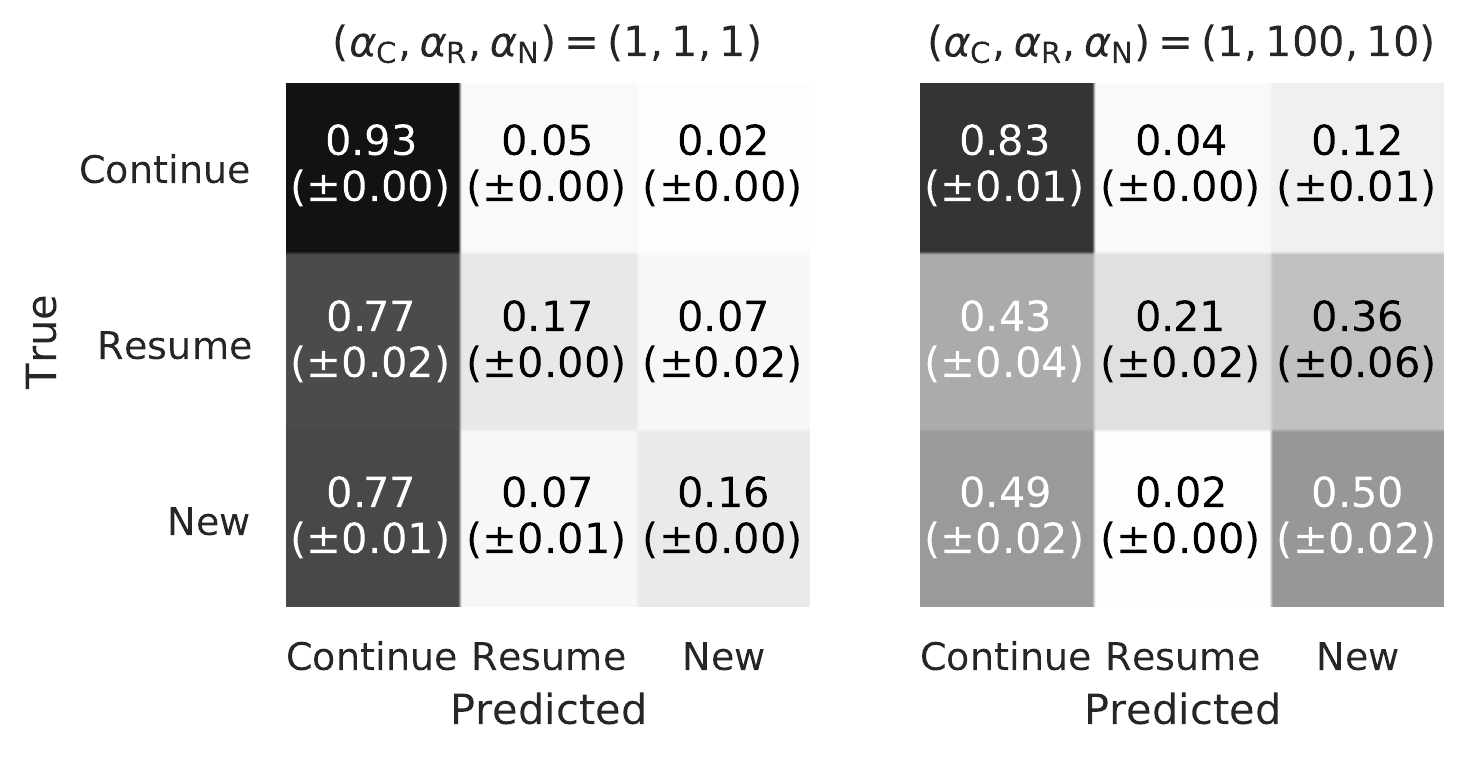}
  \caption{Scenario confusion matrix as we vary the scenario's weight $\alpha_s$ used in the loss when using teacher forced history.}
  \label{fig:teacher-forced-thread-history-confusion-matrices}  
\end{minipage}
\hspace{0.02\linewidth}
\begin{minipage}{0.49\linewidth}
  \includegraphics[width=1\linewidth]{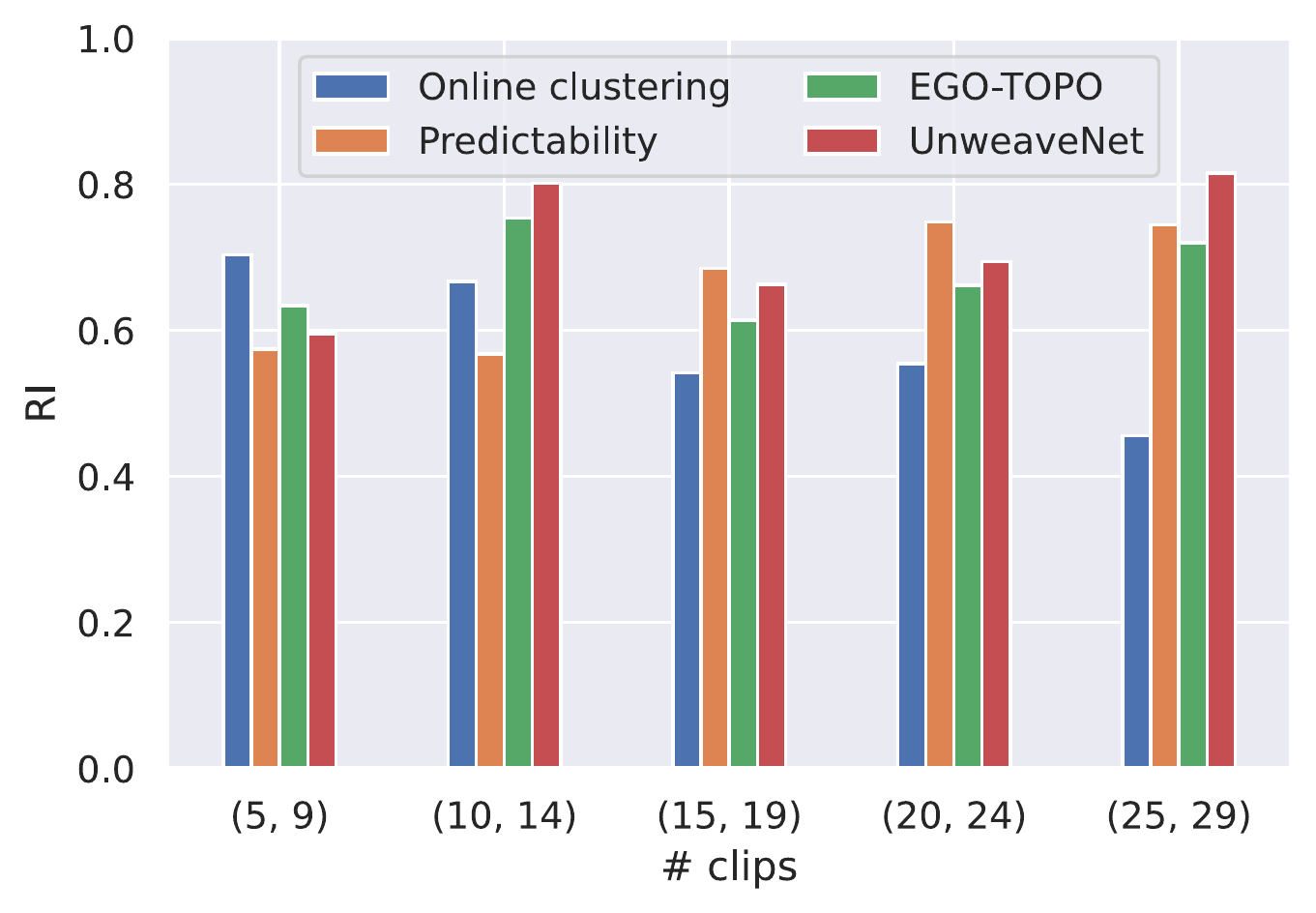}
  \caption{RI for binned stories by number of clips. Plot highlights UnweaveNet is robust to length of stories.}
  \label{fig:binnedPerClips}
\end{minipage}
\end{figure}

\paragraph{How do the scenario loss weights affect UnweaveNet's behaviour?}
\Cref{chap:unweavenet:sec:method:training} noted that it was necessary to weight different scenarios in the loss function.
The effects of the scenario loss weights $\alpha_s$ used in \cref{eq:loss} on the model's behaviour are investigated in \cref{fig:teacher-forced-thread-history-confusion-matrices}.
When an equal weighting for all scenarios is used (left), the model is heavily biased towards continuing threads.
To mitigate this, a higher weighting $(\alpha_\mathrm{C}, \alpha_\mathrm{R}, \alpha_\mathrm{N}) = (1, 100, 10)$ is used for resume and new thread scenarios (right) roughly proportional to their inverse frequency in the training split.
The higher weighting for these scenarios improves their performance at the cost of the continue thread scenario.
As anticipated, resuming threads after a break is the hardest scenario to tackle.
Two additional configurations for $(\alpha_\mathrm{C}, \alpha_\mathrm{R}, \alpha_\mathrm{N})$ are shown in the center of the figure between the two extremes of an equal weighting for each scenario (left) and heavily weighting towards resume and new thread scenarios (right).
Overall, the proposed non-uniform weightings perform better in terms of RI (75.1\%) than using equal scenario weightings (67.9\%).

\paragraph{How does the performance vary for different story lengths?}
To showcase that Unweavenet is robust to story length, we analyse the results for stories of varying lengths.
In Fig~\ref{fig:binnedPerClips}, we bin stories by \# of clips in the story and compare RI for each bin across the baselines and UnweaveNet.
Results demonstrate that UnweaveNet is robust to story length whereas online clustering struggles.

\end{document}